\documentclass[lettersize,journal]{IEEEtran}

\usepackage{amsmath,amsfonts}
\usepackage{algorithmic}
\usepackage{algorithm}
\usepackage{array}
\usepackage[caption=false,font=normalsize,labelfont=sf,textfont=sf]{subfig}
\usepackage{textcomp}
\usepackage{stfloats}
\usepackage{url}
\usepackage{verbatim}
\usepackage{graphicx}
\usepackage{cite}
\usepackage[colorlinks,urlcolor=blue,linkcolor=red,citecolor=blue]{hyperref}

\usepackage[dvipsnames]{xcolor}

\newcommand{\alphav}{\boldsymbol{\alpha}}

\begin{document}

\title{QuantNAS for super resolution: searching for efficient quantization-friendly
architectures against quantization noise}

\author{ Egor Shvetsov$^\dagger$\thanks{$^\dagger$ indicates equal contribution.},
Skoltech.
%Moscow, Russia\\
% {\tt\small e.shvetsov@skoltech.ru}
\and
Dmitry Osin$^\dagger$, Skoltech.
%Moscow, Russia\\
% {\tt\small d.osin@skoltech.ru}
\and
Ivan Koryakovskiy, Yandex.
%Moscow, Russia\\
%{\tt\small  i.koryakovskiy@gmail.com}
\and
Valentin Buchnev, Yandex.
%Moscow, Russia\\
%{\tt\small  buchnev.valentin@gmail.com}
\and
Evgeny Burnaev, Skoltech.
\and
Alexey Zaytsev, Skoltech.}
%Moscow, Russia\\
% {\tt\small   e.burnaev@skoltech.ru}

        % <-this % stops a space

\markboth{Journal of \LaTeX\ Class Files,~Vol.~14, No.~8, August~2021}%
{Shell \MakeLowercase{\textit{et al.}}: A Sample Article Using IEEEtran.cls for IEEE Journals}


\maketitle

\begin{abstract}
    % \ik{Is it possible to proofread the text? If not, we can try ChatGPT to polish the text. I feel that it is needed.}
   There is a constant need for high-performing and computationally efficient neural network models for image super-resolution%
   % \ik{Avoid acronyms in abstract. "acro" package can help to automatically track acronyms.}
   : computationally efficient models can be used via low-capacity devices and reduce carbon footprints. 
   One way to obtain such models is to compress models, e.g. quantization. 
   Another way is a neural architecture search that automatically discovers new, more efficient solutions. We propose a novel quantization-aware procedure, the QuantNAS that combines pros of these two approaches.
   
   To make QuantNAS work, the procedure looks for quantization-friendly super-resolution models. The approach utilizes entropy regularization, quantization noise, and Adaptive Deviation for Quantization (ADQ) module to enhance the search procedure. The entropy regularization technique prioritizes a single operation within each block of the search space. Adding quantization noise to parameters and activations approximates model degradation after quantization, resulting in a more quantization-friendly architectures. ADQ helps to alleviate problems caused by Batch Norm blocks in super-resolution models.

Our experimental results show that the proposed approximations are better for search procedure than direct model quantization. 

QuantNAS discovers architectures with better PSNR/BitOps trade-off than uniform or mixed precision quantization of fixed architectures.  We showcase the effectiveness of our method through its application to two search spaces inspired by the state-of-the-art SR models and RFDN. Thus, anyone can design a proper search space based on an existing architecture and apply our method to obtain better quality and efficiency.
 
% We compare our results with existing state-of-the-art approaches like EdMIPS and RFDN. 

The proposed procedure is 30\% faster than direct weight quantization and is more stable.
\end{abstract}


\section{Introduction}

\IEEEPARstart{N}{eural} networks (NNs) have become a default solution for many problems because of their superior performance. 
However, wider adoption of NNs is often hindered by their high computational complexity, which poses challenges, particularly for mobile devices. Ensuring computational efficiency is crucial, especially in tasks like super-resolution~\cite{anwar2020deep}, where deep learning models are employed to enhance image quality.

The reduction in model size not only saves costs for companies that frequently use large models but also contributes to addressing climate change by reducing the carbon footprint associated with model training ~\cite{xu2023survey}.

General and domain-specific~\cite{song2020superpet} models appear in these domain.
Modern SOTA approaches~\cite{RFDN} include many heavy blocks intended for increasing quality of images, rapidly improving existing works~\cite{romano2016raisr}.

Researchers try to reduce the complexity of NNs via compression, search, or combination of these approaches~\cite{review}.
During \emph{compression}, we try to imitate a bigger model with a smaller alternative.
A quantization of models' parameters is a vital approach of this direction~\cite{OQAT, HWGQ, LSQ, QAT, DiffQ, PQN}, as it directly reduces bit width for each parameter reducing the model size and inference time.
During \emph{search}, we look for new more efficient structures.
Searches are often done via Neural Architecture Search (NAS) \cite{AGD, Darts, DARTSPT, edmips},
% , FBNet, ISTA, BATS, Trilevel}, 
where we do structural optimization in some search space of architectures.

The quantization is a non-trivial operation.
Straightforward reduction of bit-length for the storage of a single parameter leads a to significant decrease of the model quality.
So, the models are trained via Quantization-aware-training.
A quantization is a non-differentiable operation, while 
some solutions were proposed in \cite{STE, DiffQ}, relaxing a non-differentiable optimization problem to a close differentiable one.
However, optimizing quantized weights still tends to take longer to converge, and may result in sub-optimal solutions.

NAS needs to run search in a discrete space of possible neural network architectures. 
DARTS~\cite{Darts} introduces a continuous relaxation of the discrete architecture choices and formulates the search problem as an optimization task. By using the relaxation, the search space becomes differentiable, enabling the use of gradient-based optimization algorithms. 
This relaxation is achieved via supernet construction.
By selecting a part of a supernet, we obtain a separate neural network for the problem at hand.
Discrete choices are transforming into a weighted sum of possible paths, thus creating a large network that encompasses all possible architectures.

Differentiable NAS has shown success in searching for efficient architectures while considering hardware constraints. 
Methods in \cite{FBNet, OQAT, edmips} use differentiable NAS to estimate architecture performance by examining coefficients in a weighted sum of operations within layers of a supernet. 

Authors in AGD \cite{AGD} and TrilevelNAS \cite{Trilevel} have applied differentiable NAS methods to search for super-resolution (SR) architectures. TrilevelNAS, in particular, focuses on developing computationally efficient architectures by introducing a new search space and proposing a novel search procedure. While their method shows promising results, it is still time-consuming, and they do not consider further models quantization, which limits its practical application.

The combination of NAS and quantization techniques is an even more difficult problem, as we jointly search for efficient quantization-friendly models for SR.  A natural way to approach this problem is to expand the search space by including identical operations with different low bit values.

The OQAT approach \cite{OQAT} explored quantization-friendly architectures through NAS, but was limited by the use of uniform quantization with fixed bit values. On the other hand, BOMP-NAS \cite{van2023bomp} combined mixed-precision quantization and NAS, but its application was restricted to image classification tasks on CIFAR10 and CIFAR100 datasets. While these approaches provide valuable insights into the integration of mixed-precision quantization and NAS, their applicability to the specific domain of SR remains to be explored. 

% \ik{Can we try BOMP-NAS on SR? Currently, text reads as if it could our main comparison benchmark.}

Straightforward combination of NAS and mixed-precision quantization leads to unstable and slow convergence caused by the search space size and non-differentiable quantization operations. 
% \ik{It would be nice to prove it, e.g., by using BOMP-NAS on SR} 
These problems will be amplified even further, as Batch Norm (BN) in SR models can negatively impact final performance and is usually removed from SR architectures \cite{AdaDM,Esrgan, RFDN, review}, training models without BN significantly slow down convergence. 

This work proposes a solution to the NAS for SR models that handle these problems. 
The contribution of our article are the following:
\begin{itemize}
\item We introduce a novel Neural Architecture Search (NAS) approach for mixed precision quantization architectures intended for SR that we call QuantNAS. 
It efficiently searches for low-resource architectures.
\item Innovations of our approach include entropy regularization, Search Against Noise (SAN) technique, and Adaptive Deviation for Quantization (ADQ) module. These enhancements improve stability, speed, and overall performance of the result, making possible the observed empirical improvements.
\item  Within QuantNAS, we propose an SR-friendly search space. The design is informed by analysis of efficient SR models, allowing us to adapt our approach effectively. Additionally, we showcase the effectiveness of our method through its application to a search space inspired by the state-of-the-art SR model RFDN \cite{RFDN}. 
\item In both experimental settings, we provide evidence supporting the benefits of employing NAS with mixed precision quantization, in contrast to solely using NAS or mixed precision quantization for fixed models. Thus, one can design a proper search based on an existing architecture to obtain better quality and efficiency.
\item Our joint NAS and quantization procedure yields superior Pareto fronts compared to individual NAS or mixed-precision quantization approaches.
\end{itemize}

\section{Related works}

Given the diversity of possible solutions and widespread adaption of continuous optimization approach, the most common approach to neural architecture search is based on relaxation of this problem to a differentiable one and solution of a relaxed problem~\cite{Darts}.
In this section, we start with an overview of differentiable NAS.
Then, we consider works that are more related to the search of quantized architectures via differentiable NAS.  
Finally, we focus on two specific parts of any NAS algorithm: optimization approach and search space, and how one should approach such a problem for quantization-aware mixed-precision NAS.

\textbf{Differentiable NAS (DNAS)}~\cite{FBNet, AGD, Trilevel} is a differentiable method of selecting a directed acyclic sub-graph (DAG) from an over-parameterized supernet.
Supernet includes all possible variations of architecture that we aim to select from.
Specifically, it consists of a number of layers, for each of which we have a set of nodes such that each node corresponds to a specific operation. Output of a layer is a weighted sum of nodes within this layer. Weights used in such operation are called importance weights.

% each node is summed with some importance weight representing weighted edge. 
% FIXED: IMPORTANCE WEIGHTS ARE NOT DEFINED BEFORE, READER DON'T KNOW WHAT THEY ARE!

\begin{figure}[ht]
\begin{center}
{\includegraphics[scale=0.3]{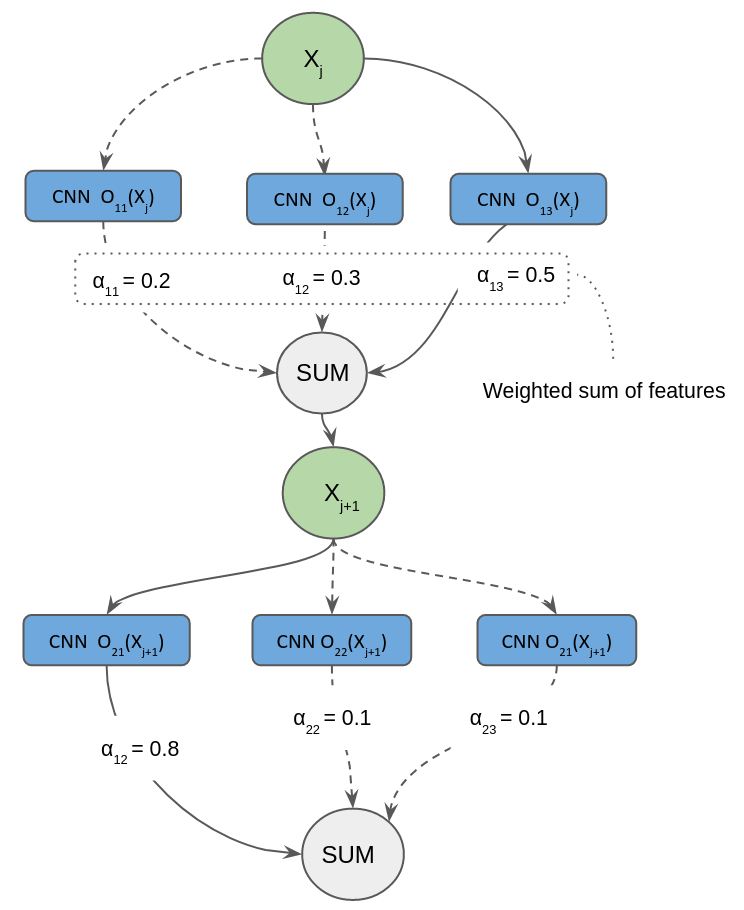}}
 \end{center}
\caption{The example of an overparametrized search space suitable for NAS. An overparametrized supernet is a graph. In this graph, multiple possible operation edges connect nodes that are outputs of each layer. The $\alpha$ values represent the edge importance. The joint training of operation parameters and their importance allow for differentiable NAS. The final architecture is the result of the selection of edges with the highest importance between each consecutive pair of nodes. The selected edges are marked with solid lines, composing a final neural network architecture.}
\label{fig:dag_supernet}
\end{figure}

% By selecting a single path from a supernet via choosing most important edge at each layer, we obtain a final architecture.  
During the search procedure, we aim to assign importance weights for each edge and consequently select a sub-graph using edges with the highest importance weights.
An example of such selection is in Figure~\ref{fig:dag_supernet}.

% DON'T BE CRUEL TO YOUR READER!

%    , ,, ,                              
%    | || |    ,/  _____  \.             
%    \_||_/    ||_/     \_||             
%      ||       \_| . . |_/              
%      ||         |  L  |                
%     ,||         |`==='|                
%     |>|      ___`>  -<'___             
%     |>|\    /             \            
%     \>| \  /  ,    .    .  |           
%      ||  \/  /| .  |  . |  |           
%      ||\  ` / | ___|___ |  |     (     
%   (( || `--'  | _______ |  |     ))  ( 
% (  )\|| (  )\ | - --- - | -| (  ( \  ))
% (\/  || ))/ ( | -- - -- |  | )) )  \(( 
%  ( ()||((( ())|         |  |( (( () )

The weights assignment can be done in several ways. The main idea of DNAS is to update importance weights $\alphav$ with respect to a loss function parameterized on supernet weights $W$.

DNAS has been proven to be efficient to search for computationally-optimized models. 
In this case, hardware constraints are introduced as an extension of an initial loss function.
FBnet~\cite{FBNet} focuses on optimizing FLOPs and latency with the main focus on classification problems. 
AGD~\cite{AGD} and TrilevelNAS~\cite{Trilevel} apply resource constrained NAS for super resolution problem (SR) by minimizing FLOPs during search procedure.

\begin{figure}[h]
\begin{center}
{\includegraphics[scale=0.3]{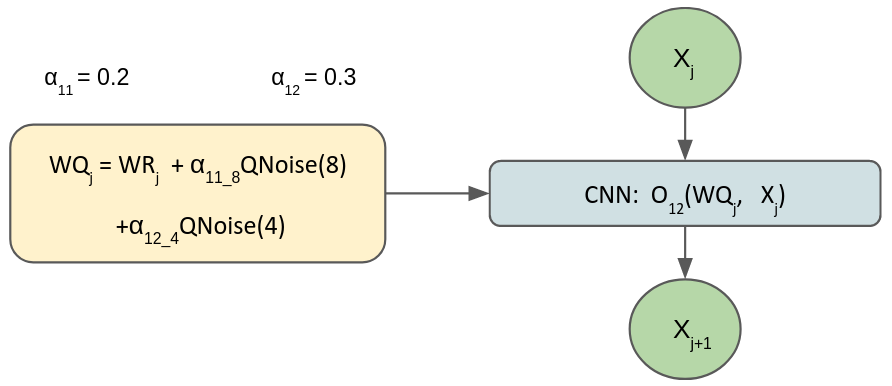}}
 \end{center}
\caption{SAN approach for a single layer 
% \ik{Note that it is better to use proper English, i.e. to use both a subject and a verb.}. 
A function $QNoise(b)$ generates quantization noise. $WR$ are real valued weights,  $WQ$ are output pseudo quantized weights, and $\alphav$ is a vector of trainable parameters. 
By adjusting $\alphav$, we search for acceptable model degradation caused by quantization procedure.
$QNoise(b)$ is independent of weights and allows for propagation of gradients. 
For quantization-aware search, each blue operation on Figure \ref{fig:dag_supernet} becomes SAN operation with noisy weights. }
\label{fig:dag_san}
\end{figure}

\textbf{Quantization-aware DNAS.}  DNAS can be employed to search for architectures with desired properties. In OQAT \cite{OQAT}, authors performed a search for architectures that perform well when quantized. They specifically used uniform quantization, where the same number of bits is used for each layer of the neural network. The architectures discovered through quantization-aware search performed better when quantized compared to architectures found without considering quantization.

Uniform quantization has limitations in terms of flexibility compared to mixed-precision quantization (MPQ). In MPQ, each operation and activation in the neural network has its own bit value. 
This allows for more fine-grained control over the precision of different parts of the network, potentially leading to better performance. In EdMIPS~\cite{edmips}, authors focus on finding an optimal allocation of quantization bit values for each layer via differential NAS like procedure.  

% \ik{I would appreciate if you also cite our Huawei paper "One-Shot Model for Mixed-Precision Quantization" :)}

The use of quantization techniques like Straight-Through Estimator (STE) \cite{STE} can introduce oscillations during optimization due to rounding errors. DiffQ \cite{DiffQ} addresses this issue by introducing differentiable Quantization Noise (QN) to approximate the degradation caused by quantization. Notably, DiffQ only applies QN to model weights. In NIPQ~\cite{shin2023nipq}, authors combine QN and LSQ (Learned Step Size Quantization) \cite{LSQ} by sharing the same parameter corresponding to the same bit levels. This approach facilitates an easy transition from QN to LSQ during the later stages of training, allowing for improved quantization performance.

\textbf{Efficient Super Resolution architectures.}
Many current models for SR suffer from high computational costs, making them impractical for resource-constrained devices and applications. To address this issue, lightweight SR networks have been proposed. One such network is the Information Distillation Network (IDN)~\cite{hui2018fast}, which splits features and processes them separately. Inspired by IDN and IMDB~\cite{hui2019lightweight}, the RFDN~\cite{RFDN} improves the IMDB architecture by using RFDB blocks~\cite{RFDN}. These blocks utilize feature distillation connections and cascade 1x1 convolutions towards a final layer.

While there are numerous ideas for making SR models lightweight, developing such methods can be labor-intensive due to the trial-and-error process. In our work, we aim to improve existing architectures - specifically the RFDN network, which was the winning solution in the AIM 2020 Challenge on Efficient Super-Resolution \cite{zhang2020aim}. We focus on modifying RFDN to be more amenable to quantization by constructing a quantization-aware, RFDN-based space.

%\ik{bit width is a correct spelling (neither bitwidth nor bit width)}
\textbf{Search space design} is crucial. 
It should be both flexible and contain known best-performing solutions. 
Even a random search can be a reasonable method with a good search space design.
In AGD~\cite{AGD}, authors apply NAS for SR, and search for \textbf{(1) a cell} - a block which is repeated several times, and \textbf{(2) kernel size}, along with other hyperparameters like the number of input and output channels. 
TrilevelNAS\cite{Trilevel} extends the previous work by adding \textbf{(3) network level} that optimizes the position of the network upsampling layer. 

% In our work, we employ two strategies. Firstly, we design our own search space to explore a wide range of architectural possibilities. Secondly, we conduct a search centered around a well-performing architecture. Specifically, we select RFDN \cite{RFDN} as our base architecture due to its demonstrated success as a computationally efficient super-resolution model. 

\textbf{Supernet co-adaption during differentiable architecture search} makes it difficult to a select final architecture from the supernet because selected operations depend on all the left in the supernet operations. Therefore, we need to explicitly enforce operations independence during search phase. Below, we dicuss available solutions.
%  \cite{BATS,entropy,GUMBEL,sparse,Trilevel,ISTA}. 

Enforcing operations independence depends on the graph structure of a final model. In our work, we use the Single-Path graph - one possible edge between two nodes (more in Appendix~\ref{sec:single_path}). 
For this structure, the sum of node outputs is a weighted sum of features (see Figure \ref{fig:dag_supernet}), the co-adaptation problem becomes obvious. 
Second layer convolutions are trained on a weighted sum of features, but after selecting a subgraph via discretization, only one source of features remains. Therefore, enforced independence for the vector of $\alphav$ is necessary. 
In BATS \cite{BATS}, independence is achieved via scheduled temperature for softmax. 
Entropy regularization is proposed in Discretization-Aware search \cite{entropy}. 
In \cite{GUMBEL}, authors proposed an ensemble of Gumbels to sample sparse architectures for the Mixed-Path strategy, and in \cite{wu2021neural}, Sparse Group Lasso (SGL) regularization is used. 
In ISTA-NAS \cite{ISTA}, authors tackle sparsification as a sparse coding problem. Trilvel NAS \cite{Trilevel}  proposed sorted Sparsestmax. 

\textbf{Summary.}
Many works approached problems of NAS for fixed-bit and quantized architectures by introducing differentiable NAS and considering various search spaces.
However, there are no approaches that can efficiently solve the problem of NAS for mixed-precision quantized architectures for the SR.
This is natural, because the problem to solve is challenging due to extensive search space, unstable training, and large amount of resources required.
We believe that with a proper design of NAS, this problem can become computationally tractable and will produce new interesting architectures that are suitable for low-resource devices.

\section{Methodology}

%We follow the basic definitions provided in the previous section with the description of our approach. 
The description of the methodology consists of four parts.
We start with Subsection~\ref{sec:design} that describes our search spaces.
Subsection~\ref{sec:ADQ} describes our ADQ module, which is specifically designed to substitute Batch Norm and make the search space more robust.
In Subsection~\ref{sec:quant} we introduce mixed precision search and provide details on Search Against Noise technique.
The complete QuantNAS search procedure is described in subsection~\ref{sec:train_search}. 
It includes the description of the used loss function.

\subsection{Search space design}
\label{sec:design}

\begin{figure*}[t]
\center{\includegraphics[scale=0.35]{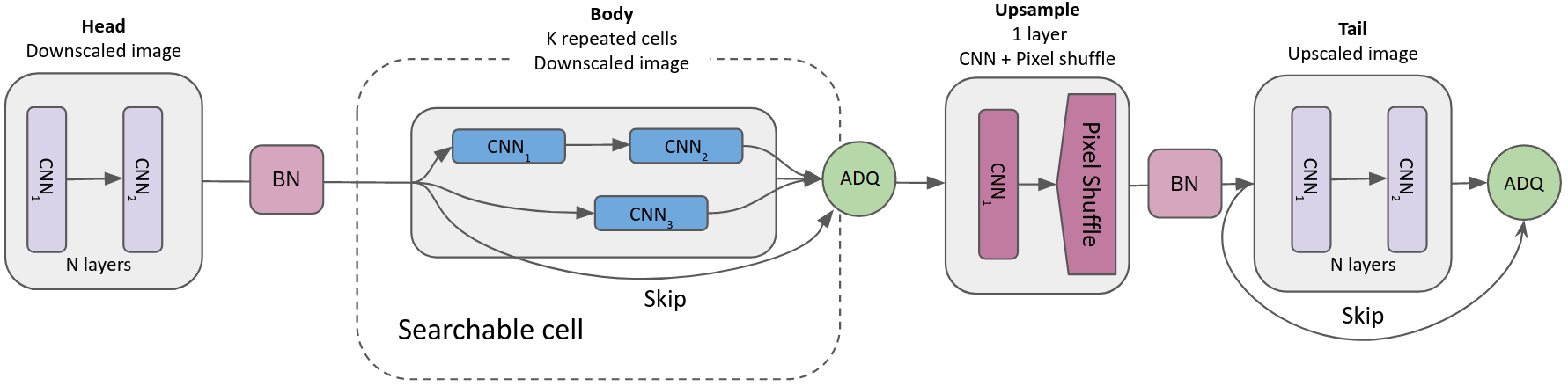}}
\caption{The search space design. We separate the whole architecture into $4$ parts: head, body, upsample, and tail. The head and the tail have $N = 2$ convolutional layers. The identical body part is repeated $K = 3$ times, unless specified otherwise. The number of channels for all the blocks equals $36$, except for the head's first layer, upsample, and the tail's first layers. All the blocks with skip connections incorporate ADQ.}
\label{fig:search_design}
\end{figure*}

The work considers two approaches to design the search space.
For designing the first search space, which we call \emph{Basic search space}, we take into account recent results in this area and, in particular, the SR quantization challenge~\cite{SRContest}. 
We combine most of these ideas in the search design depicted in Figure~\ref{fig:search_design}. 
The second search space RFDN expands a recent computationally efficient architecture RFDN~\cite{RFDN}.

\emph{Basic search space} consists of head, body, upsample, and tail blocks. The \textbf{head} block is composed of two searchable convolutional layers. These layers play a crucial role in the model and are responsible for capturing important features at the beginning of the network.  The \textbf{body} block consists of three layers. It includes two consecutive layers and one parallel layer, along with a skip connection. This block can be repeated multiple times to enhance the model's performance. Each body block is followed by ADQ. The \textbf{upsample} block consists of one searchable convolutional layer and one upsampling layer. The upsampling operation is performed using the Pixel Shuffle technique, as described in ESPCN~\cite{ESPCN}. This block is responsible for increasing the resolution of the image. The \textbf{tail} block consists of two searchable convolutional layers with a skip connection. This block is located at the end of the network and is responsible for refining the features and generating the final output. \emph{Basic search space} is depicted in Figure~\ref{fig:search_design}.

\textbf{The deterministic part }of our search space includes the position of upsample block and the number of channels in convolutions.
The ADQ block is used only in quantization-aware search. \textbf{The variable part} refers to quantization bit values and operations within head, body, upsample, and tail blocks.  
All possible operations are defined in Appendix section~\ref{sec:q_space}.
We perform all experiments with 3 body blocks, unless specified otherwise. 

To create the \textit{RFDN search space}, we start with the RFDN architecture and replace all convolutional layers with searchable operations. The possible operations are listed in the Appendix, specifically in section~\ref{sec:q_space}. Each operation has different bit values that can be searched. The key difference between the \textit{Basic search space} and the \textit{RFDN search space} is that the latter uses repeated RFDB blocks~\cite{RFDN} instead of body blocks defined above, the tail block has only 1 layer, and there is no head block. Instead, 3 input channels are repeated and concatenated to have a desirable shape for RFDB block. 

If we were to substitute the body block in the \textit{Basic search space}  with the RFDB block, it would result in an architecture very similar to RFDN~\cite{RFDN}. The \textit{Basic search space} can be easily modified to create various popular SR architectures by adjusting the structure of the inner blocks, which is why it is called the \textit{Basic search space}.

% Empirically, we found that the original solution with two BN layers did not improve the quality for quantized models. However, the same approach with only one BN layer improved performance, quantitative results for different modifications of ADQ can be seen in Table~\ref{tab:ADQ}.

\subsection{ADQ module}
\label{sec:ADQ}
% \egor{Describe why we need this module and the problem of BN layers in SR models. An idea that BN may improve overall search procedure.}

Variation in a signal is crucial for identifying small details for the SR preventing usage of normalizations like batch norm (BN).
After normalization layers, the residual feature's standard deviation shrinks, causing the performance degradation in SR task~\cite{AdaDM}.
On the other hand, training a neural network without BN is unstable and requires more iterations.
The issue is even more severe for differentiable NAS, as it requires training an overparameterized supernet.

\begin{figure}[t]
\center{\includegraphics[width=0.45\textwidth]{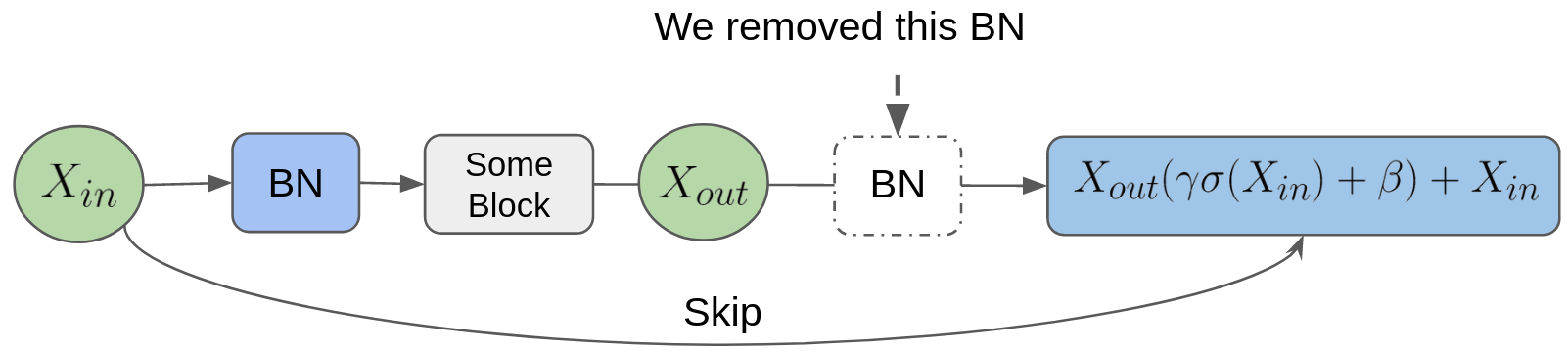}}\caption{Comparison of ADQ with  AdaDM~\cite{AdaDM}. \emph{Some Block} represents any residual block with several layers within, $\sigma(X_{in})$ is a variance of input signal, $\gamma$ and $\beta$ are learnable scalars. We remove the second BN after $X_{out}$ from original AdaDM.}
\label{fig:ADM}
\end{figure}

The authors of AdaDM~\cite{AdaDM} proposed to rescale the signal after BN, based on its variation before BN layers. 
We empirically proved that removing the second BN in AdaDM scheme, keeping only the first one in each block, leads to better results for quantized models. We call this block ADQ. 
Original AdaDM block and our modification are depicted in Figure~\ref{fig:ADM}. 
All the body blocks during search have the ADQ module.
% the body block, all the repeated layers within the body block and the tail block, 

\subsection{Quantization-Aware Training - QAT}
\label{sec:quant}

Our aim is to find quantization-friendly architectures that perform well after quantization. 
A standard approach to obtain a trained and quantized model is the Quantization-Aware Training \cite{QAT}. For QAT, we sequentially perform the following: (a)~quantize full precision weights and activations during forward pass; (b)~compute gradients using STE~\cite{STE} by bypassing non differentiable quantization operation; and (c)~update full precision weights. 
% that give us an efficient and high-performing procedure QuantNAS.
\label{sec:final_quantization}

Let consider the following one-layer neural network (NN) with input $x$,
\begin{equation} 
\label{eq:nn_simple}
y = f(a(x)) = W a(x),
\end{equation}
where $a$ is a non linear activation function and $f$ is a function parametrized by a tensor of parameters $W$. 
While in~\eqref{eq:nn_simple} $f$ is a linear function, a convolutional operation is also a linear function, so the structure is general enough. 
To decrease the computational complexity of the network, we replace expensive float-point operations with quantized operations. Quantization occurs for both weights $W$ and activation $a$. 

The quantized output has the following form:
\begin{equation} 
\label{eq:quant_simple}
y_q = f_q(a_q(x)) = o(G(x, b), Q(W, b)),
\end{equation}
 where quantization bit width is denoted as $b$ and a convolution layer is denoted as $o$.
 
 $Q(W, b)$ is a quantization function for weights.
 We use Learned Step Quantization (LSQ)~\cite{LSQ} with trainable step value. 
 
 $G(W, b)$ is a quantization function for activations.
 We use a half wave Gaussian quantization function~\cite{HWGQ} for it.  

\subsubsection{Mixed precision Search and BitMixer}
\label{sec:bitmixer}
The task of mixed-precision quantization is to find optimal bit width for each layer in a neural network. In this scenario, we replace each convolution layer with an operation that we call BitMixer. BitMixer's purpose is to model a weighted sum of the same convolutional operation quantized to different bit width during search.

The straightforward approach is to have an independent set of weights for each convolutional operation. Let $\alphav$ be vector of importance weights corresponding to different bit width. Then, for convolution $o$ and input $x_l$, the output of $l$-th layer is:

\begin{equation} 
\label{eq:super_net_quant_basic}
    \begin{split}
     BitMixer(\alphav, o, x_l) =  \sum_{b \in B} \alpha_{b} \cdot o \biggl( G\bigl(x_l, b\bigl),  Q(W^o_b, b) \biggl)
    \end{split}
\end{equation}
This approach requires computing the same convolutional operation $|B|$ times.
\subsubsection{Quantization-Aware Search with Shared Weights (SW)} 
To improve computational efficiency we can quantize weights of identical operations with different quantization bits instead of using different weights for each quantization bit, this idea was studied in~\cite{edmips}. Then \eqref{eq:super_net_quant_basic} becomes:

\begin{equation} 
\label{eq:super_net_quant_shared}
    \begin{split}
     BitMixer(\alphav, o, x_l) =  
      \biggl(\sum_{b \in B}{\alpha_b}\biggl) \cdot \\ \cdot o \biggl( \sum_{b \in B} \hat{\alpha}_{b} G\bigl(x_l, b\bigl),  \sum_{b \in B} \hat{\alpha}_{b} Q(W^o, b) \biggl),
    \end{split}
\end{equation}
where $ \hat{\alpha}_b = \frac{\alpha_b}{\sum_{b \in B} \alpha_b}$. 

Note that it is not necessary to use the first term of the product as well as alpha-scale  when $\sum_{b \in B}{\alpha_b} = 1$. This is the case when we only try to find optimal bit-width for a layer but \textbf{do not} search for convolutional operation, like in~\cite{edmips}. 
QuantNAS, however, searches for different bit width and operation simultaneously, which is why we perform such adjustments. Without it, $\sum_{b \in B}{\alpha_b}$ is significantly smaller than 1, with forward signal magnitude being drastically reduced after going through BitMixer.

% \begin{equation} 
% \label{eq:super_net_quant_shared}
%  x_{j+1} = \sum_{i=1}^{|O^j|} o_{ij} \left( \sum_{b \in B} \alpha_{jib} G\bigl(a(x_j), b\bigl), \sum_{b \in B} \alpha_{jib} Q(W_{ji}, b) \right).
% \end{equation}

The effectiveness of SW can be seen from~\eqref{eq:super_net_quant_shared}: it requires fewer convolutional operations and less memory to store the weights.

% But gradients with respect to weights and quantization functions still should be computed for each new added bit value.

\subsubsection{Quantization-Aware Search Against Noise (SAN)}

To further improve computational efficiency and performance of search phase, we introduce SAN.
Model degradation caused by weights quantization is equivalent to adding the quantization noise
$QNoise_b(W) = Q(W, b) - W$.
Then, quantized weights is $Q(W, b) = W + QNoise_b(W)$ and \eqref{eq:super_net_quant_shared} is:

\begin{equation} 
 \label{eq:super_net_quant_noise}
 \begin{split}
     &BitMixer(\alphav, o, x_l) =  
      \biggl(\sum_{b \in B}{\alpha_b}\biggl) \cdot \\ \cdot o \biggl(\sum_{b \in B} \hat{\alpha}_{b} &QNoise_{b}\bigl(x_l\bigl) + x_l, \sum_{b \in B} \hat{\alpha}_{b} QNoise_{b}\bigl(W^o\bigl) + W^o\biggl)
 \end{split}
\end{equation}

% \begin{equation} 
%  \label{eq:super_net_quant_noise}
%  \begin{split}
%  x_{j+1} = \sum_{i=1}^{|O^j|} o_{ij}\biggl(&\sum_{b \in B} \alpha_{jib} QNoise_{b}\bigl(a(x_j)\bigl) + a(x_j), \\
%                 &\sum_{b \in B} \alpha_{jib} QNoise_{b}\bigl(W_{ji}\bigl) + W_{ji}\biggl).
%  \end{split}
% \end{equation}
This procedure does not require weights quantization and is differentiable, unlike straightforward quantization. 
$QNoise_b$ is a function of $W$ because it depends on its shape and magnitude of values. 
Given the quantization noise, we can more efficiently run forward and backward passes for our network, similar to the reparametrization trick. 

% After searching with such a procedure, we obtain a quantization-friendly architecture with optimal quantization bits for each operation.

Adding quantization noise is similar to adding independent uniform variables from $[- \Delta/2, \Delta/2]$ with $\Delta = \frac{1}{2^b-1}$, as explained in in~\cite{PQN}.
However, for the noise sampling, we use the following procedure as in ~\cite{DiffQ}:
\begin{equation} \label{eq:noise}
QNoise(b) = \frac{\Delta}{2} z, z \sim \mathcal{N}(0,1),
\end{equation}
as it performs slightly better than the uniform distribution~\cite{DiffQ}.

\subsection{The search procedure}
\label{sec:train_search}
The search and training procedures are carried out as two separate steps. First, we search for an architecture and bit width, and then we conduct another training session for the selected architecture. We assign individual  $\alpha$-importance values to each possible operation with a specific bit. This means that the number of  $\alpha$ values is equal to the number of operations multiplied by the number of possible bits. For $l$-th layer, there are $|O_l|$ possible operations and $|B|$ bit widths.

For search step, we alternately update supernet's weights $W$ and edge importances $\alphav$. 
Two different subsets of training data are used to calculate the loss function and derivatives for updating $W$ and $\alphav$, similar to~\cite{AGD}.
Hardware constraints and entropy regularisation are applied as additional terms in the loss function for updating $\alphav$. 

To calculate the output of $l$-th layer $x_{l + 1}$ we sum the outputs of BitMixer taking as inputs: importance values $\alphav_{i}^l$, convolutional operation $o_i^l$, and input $x_l$. 

% of weight the output of separate edges according to importance values $\alpha_{ib}^l$ of each of $O^l$ operations and each of $B$ bits:

\begin{equation}
\label{eq:super_net}
%\begin{align*} 
x_{l+1} = \sum_{i=1}^{|O^l|} BitMixer (\alphav_{i}^l, o^l_{i}, x_l),
%\end{align*}
\end{equation}
where $\sum_{i = 1}^{|O^l|} \sum_{b=1}^{|B|}  \alpha_{ib}^l = 1 $ and all \(\alpha_{ib}^l \geq 0\).

Note that when $|B| = 1$, $\hat{\alpha}_{b}$ used in \eqref{eq:super_net_quant_shared} and \eqref{eq:super_net_quant_noise} becomes 1, and \eqref{eq:super_net} will give us the standard DNAS procedure for searching operations.

% \begin{equation}
% \label{eq:super_net}
% %\begin{align*} 
% x_{l+1} = \sum_{i=1}^{|O^l|} \sum_{b=1}^{|B|} \alpha_{ibl}o^l_{i}(x_l),
% %\end{align*}
% \end{equation}
% where $\sum_{i = 1}^{|O^l|} \sum_{b=1}^{|B|}  \alpha_{ibl} = 1 $ and all \(\alpha_{ibl} \geq 0\).

Then, the final architecture is derived by choosing a single operator with the maximal $\alpha_{ib}^l$ among the ones for this layer. Finally, we train the obtained architecture from scratch. 

\label{sec:alpha_loss}
To optimize $\alphav$, we compute the following loss that consists of three terms:
$$
    L(\alphav) = L_1(\alphav) + \eta L_{cq}(\alphav) + \mu(t) L_{e}(\alphav),
$$ 
where $\eta$ and $\mu(t)$ are regularization constants. 
$\mu(t)$ increases with each iteration $t$, details are covered in Appendix section~\ref{sec:entropy}.
$L_1(\alphav)$ is the $l_1$-distance between high resolution and restored images averaged over a batch. $L_{cq}(\alphav)$ is the hardware constraint and $L_{e}(\alphav)$ is the entropy loss that enforces sparsity of the vector $\alphav$. The last two losses are defined in two subsections below.

\subsubsection{Hardware constraint regularization}

The hardware constraint is proportional to the number of floating point operations FLOPs for full precision models and the number of quantized operations BitOps for mixed-precision models. 
$F_{fp}(o, x)$ is the function computing FLOPs value based on the input image size $x$ and the properties of a convolutional layer $o$: kernel size, number of channels, stride, and the number of groups. We use the same number of bits for weights and activations in our setup. Therefore, BitOps can be computed as $F_{q}(o, x) = b^2 F_{fp}(o, x)$, where $b$ is the number of bits. 
Then, the corresponding hardware part of the loss $L_{cq}$ is:
\begin{equation}\label{eq:bitops}
L_{cq}(\alphav) = \sum_{l=1}^{|S|} \sum_{i=1}^{|O^l|} \sum_{b=1}^{|B|} \alpha_{ib}^l b^2 F_{fp}(o^l_{i}, x_l), 
%\\ \quad \textrm{where} \quad \sum_{i=1}^{|O^j|} \sum_{b=1}^{|B|} \alpha_{jib}=1  \quad \textrm{and} \quad \forall  \quad \alpha_{jib} \geq 0,
\end{equation} 
where $S$ is a supernet's block or layer consisting of several operations, the layer-wise structure is presented in Figure~\ref{fig:dag_supernet}, and $x_l$ is the input to $l$-th layer. 
We normalize $L_{cq}(\alphav)$ value by the value of this loss at initialization with the uniform assignment of $\alpha$, as the scale of the unnormalized hardware constraint reaches $10^{12}$. 

\subsubsection{Entropy regularization}
We use entropy regularization such that after the architecture search, the model keeps only one edge between two nodes, we call this procedure sparsification.
Let us denote as $\alphav_l$ all alphas that correspond to edges that connect a particular pair of nodes.
They include different operations and different bits.
At the end of the search, we want $\alphav_l$ to be a vector with one value close to $1$ and all remaining values close to $0$. 
% During our work, we tried other methods and found that entropy regularization \cite{entropy} works the best for our setting.
% Our implementation details are close to Discretization-Aware search~\cite{entropy} that leads to the following loss: 

The sparsification loss $L_{e}(\alphav)$ for $\alphav$ update step  has the following form:
\begin{equation}
 L_{e}(\alphav) = \sum_{l=1}^{|S|} H(\alphav_{l}),
\end{equation}
where $H$ is the entropy function, that we can calculate, as $\alphav$ admits interpretation as the categorical distribution.
The coefficient before this loss $\mu(t)$ depends on the training epoch $t$. 
The detailed procedure for regularization scheduling is given in Appendix~\ref{sec:entropy}.

\subsection{Summary}

% Given all the details of our solution, it is easy to get confused on what is going on.
% \ik{It is really strange to see such a phrase. If you say it, it means that the structure can be further improved and clarified. Perhaps, writing an algorithm or some plot would be a good addition to the method description.}
% To avoid this, 

We present the summary of our mixed-precision quantization NAS approach in this subsection:
\begin{itemize}
    \item We consider two search spaces that take origin from SR competition and from a recent RFDN~\cite{RFDN} architecture. To make the search procedure more stable and efficient, we use ADQ.
    \item For different edges in a single layer that have different bit values and identical operations, we share weights making training more efficient.
    \item As a loss function, we use a three-term function. The first term is a standard SR loss, the second one constrains FLOPs of a model forcing it to be more efficient, and the last one leads that importance weights converge to a single non-zero value for each layer.
    \item We perform Quantization-Aware search, so our architecture in the end would be  quantization-friendly. The idea is to substitute non-differentiable quantization with additive differentiable quantization noise. In this way, we ensure good quantization property of a final architecture.
\end{itemize}

\section{Results}

% \ik{Also the following part is usually unnecessary. If you add it, then perhaps it means that titles of each subsection are not clear enough and can be improved.}
The section is organized as follows:
\begin{itemize}
\item Initially, we provide an overview of the protocol and introduce the competitor methods. This segment also includes technical specifications for both our approach and the alternative methods.

\item We commence by conducting a comparative analysis between our approach and existing methods in the field of NAS and quantization for super-resolution.

\item To conclude, we present the findings of an ablation study, offering insights into how different contributions have improved our approach.
\end{itemize}

We provide the code for our experiments \href{https://anonymous.4open.science/r/QuanToaster/README.md}{here}.
% TODO overview of experiments

\subsection{Evaluation protocol}
\label{sec:techdetails}
The evaluation protocol follows that from~\cite{Trilevel} with DIV2K~\cite{DIV2K} the training dataset. 
The test datasets are Set14~\cite{Set14}, Set5~\cite{bevilacqua2012set5}, Urban100~\cite{Urban100}, and Manga109~\cite{fujimoto2016manga109}
The super-resolution scale is 4. 

In the main body of the paper, we present results on Set14. 
The results for other datasets are presented in Appendix. 
For training, we use RGB images. 
For PSNR score calculation, we use only the Y channel similarly to \cite{AGD,Trilevel}.  
Evaluation of FLOPs and BitOPs is done for fixed image sizes $256 \times 256$ and $32 \times 32$, respectively. 

To illustrate the effectiveness of our approach, we present the application of QuantNAS in two distinct settings. 
The first setting involves our carefully crafted custom-designed search space BasicSpace, which has been developed through a comprehensive analysis of the state-of-the-art (SOTA) architectures. 
Furthermore, we demonstrate the versatility of QuantNAS by applying it to the champion of the AIM 2020 Efficient Super-Resolution Challenge, namely RFDN~\cite{RFDN}.

\paragraph{Basic search space}
For all experiments, we consider the following setup if not stated otherwise. 
A number of body blocks is set to 3.
For quantization-aware search, we limit the number of operations to $4$ to obtain a search space of a reasonable size. 
Following others, our setup considers two options as possible quantization bits: 4 or 8 bits for activations and weights. 

% TODO: move to appendix
% For full precision search, there are ten different possible operations as candidates for a connection between two nodes.

\paragraph{RFDN search space}
In our approach, we substitute each convolutional layer within RFDN with a search block consisting of six possible operations. 
Each operation can be configured to use either 4 or 8 bits.
So, $12$ edges constitute the search block. 
Furthermore, we apply the ADQ module around each RFDN block. Notably, the ESA block remains consistently quantized to 8 bits.

\paragraph{Performance evaluation}
QuantNAS has the capability to discover models that exhibit varying levels of computational complexity and quality. By adjusting the hardware constraint regularization, we identify several distinct models. 
When these points are plotted on a graph, they form a Pareto plot, which serves as a means to assess the method's quality. Visual evaluation of such a graph can be conducted as follows: the more points situated to the left and higher up on the graph, the better the overall performance, as they have higher quality and lower complexity.

% A search space analysis and more technical details are provided in Appendix~\ref{sec:sp_analysis}. 

% \subsection{Different number of body blocks}
% \label{sec:body_blocks}
% A straightforward way to improve model performance is to increase the number of layers. We study how our method scales by performing search with a different number of body blocks: 1, 3, and 6. Three constellations are presented in Figure~\ref{fig:pareto_baseline}. We observe that increasing the number of blocks improves final performance and increases the number of BitOPs for architectures found  with the highest hardware regularization - each constellation is slightly shifted to the right.

\begin{figure}[ht]
 {\includegraphics[scale=0.40]{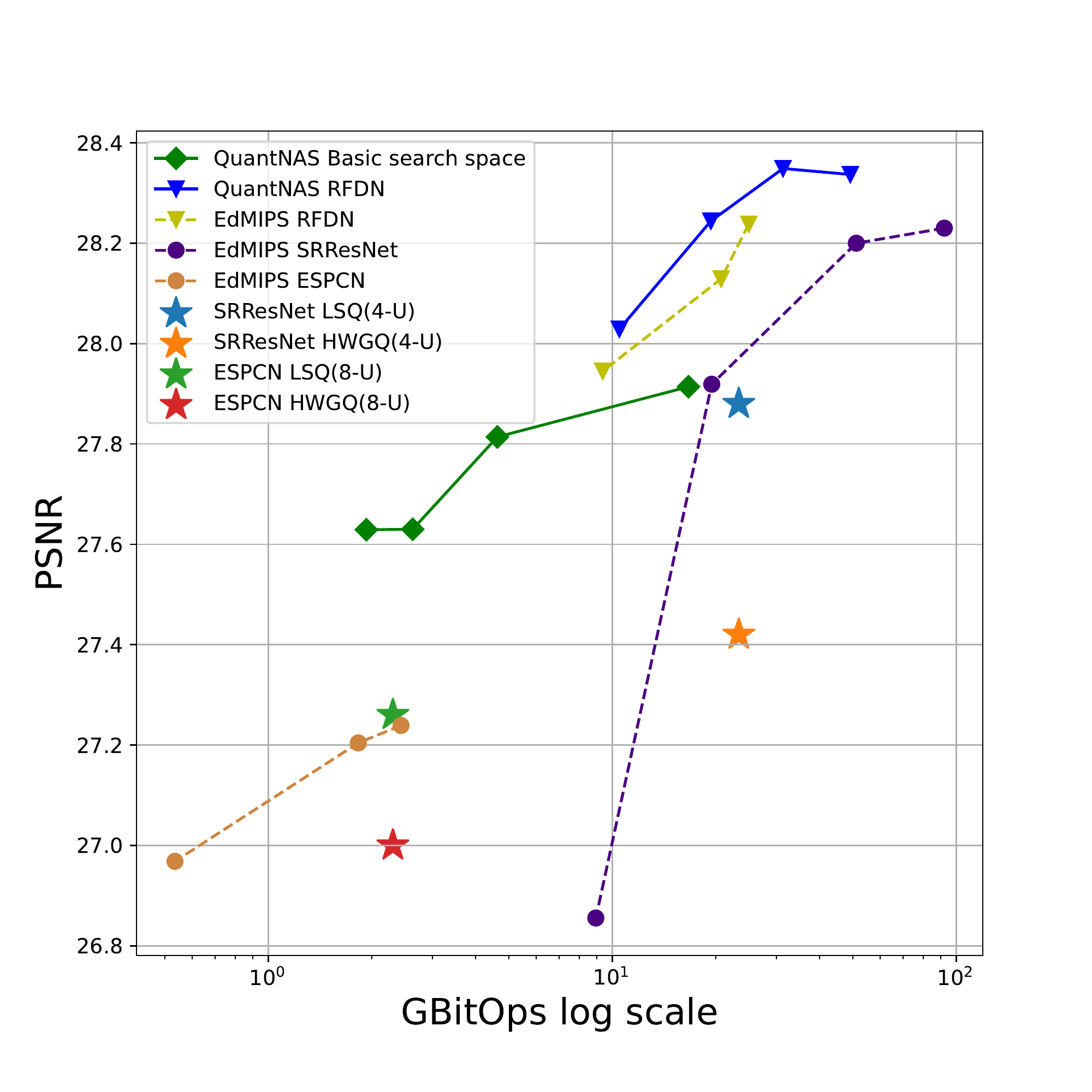}}
\caption{Our quantization-aware QuantNAS approach vs. fixed quantized architectures. PSNR is for Set14 dataset and BitOPs is for image size 32x32. We aim at the upper left corner that corresponds to smaller GBitOps and higher quality measure via PSNR.} 
\label{fig:pareto_baseline}
\end{figure}

\subsection{QuantNAS vs. quantization of fixed architectures}
\label{sec:other methods}

\paragraph{Compared methods}
To compare QuantNAS with other mixed and uniform architectures, we consider the following fixed models: SRResNet~\cite{SRRESNET}, ESPCN~\cite{ESPCN}, and RFDN~\cite{RFDN}.
For mixed precision quantization, we use EdMIPS~\cite{edmips}. 
Our setup for EdMIPS is matching the original setup and search is performed for different quantization bits for weights and activations. 
For uniform quantization, we use LSQ~\cite{LSQ} and HWGQ~\cite{HWGQ}. 

Our QuantNAS with ADQ and SAN has the following hardware penalties: 0, 1e-4, 1e-3, 5e-5 to produce distinct points at the Pareto frontier. 
Mixed precision quantization by EdMIPS \cite{edmips} for SRResNet \cite{SRRESNET}, ESPCN \cite{ESPCN}, and RFDN \cite{RFDN} used hardware penalties  0, 1e-3, 1e-2, 1e-1 respectively. 

\paragraph{Main table}
We start with comparison of different quantized models and results of QuantNAS.
ESPCN model is quantized to $8$ bits, and SRResNet is quantized to $4$ bits to match the desired model size.

Table~\ref{tab:quantized} presents the results. 
QuantNAS outputs architectures with a better PSNR/BitOps trade-off than uniform quantization techniques for both considered GBitOPs values about $5$ and about $20$. 

\paragraph{Pareto frontier}
Figure~\ref{fig:pareto_baseline} complements the table above, showcasing the complete Pareto frontier of architectures obtained using QuantNAS and EdMIPS.

QuantNAS excels in discovering architectures with more favorable PSNR/BitOps trade-offs, particularly within the range where BitOps values overlap, when compared to SRResNet and ESPCN. 
Additionally, our approach demonstrates a notable performance improvement, especially when compared to quantized ESPCN.
Moreover, it is evident that QuantNAS for RFDN delivers superior results in comparison to EdMIPS RFDN. 
% \ik{"it's" is very informal. it is better to use "it is"}

Due to computational limits, our search space is bounded in terms of the number of layers.  
We can not extend our results beyond SRResNet or RFDN in terms of BitOps to provide a more detailed comparison. 
% \ik{"can't" is also very informal. it is better to use "cannot"}

\begin{table}
\centering
\caption{Quantitative results for different quantization methods for different models. ``U'' - stands for uniform quantization - all bits are the same for all layers. GBitOPs were computed for 32x32 image size. }
% \ik{Consider, if possible, adding more previous methods here. Reviewers may complain that there are not so many methods.}
% \ik{1. Use the same number of digits after a dot 2. Why "Our search space" number is bold? It is less than EdMIPS or LSQ. Suggest to leave "RFDN search space" in bold.} 

\begin{tabular}{llll}   
    %\emph{some text}     \\ \midrule
    \hline
    Method & Model & GBitOPs  & PSNR \\ 
    \hline
     LSQ (8-U) & ESPCN  & 2.3 &  27.26     \\
    HWGQ (8-U) & ESPCN & 2.3 &  27.00     \\
     LSQ (4-U) & SRResNet  & 23.3 & 27.88   \\
    HWGQ (4-U) & SRResNet & 23.3 &  27.42       \\
    EdMIPS (4,8) & SRResNet & 19.4 &  27.92    \\
    EdMIPS (4,8) & RFDN & 20.7 & 28.13  \\
    \hline
    QuantNAS (4,8) & \textbf{RFDN} & \textbf{19.3} & \textbf{28.24}  \\
    QuantNAS (4,8) & RFDN w/o SAN & 16.8 & 28.23  \\
    QuantNAS (4,8) & RFDN w/o ADQ & 17.2 & 28.16  \\
    QuantNAS (4,8) & \textbf{Basic} & \textbf{2.6} & \textbf{27.81}   \\
    QuantNAS (4,8) & Basic w/o SAN  & 4.1 &  27.65 \\
    QuantNAS (4,8) & Basic w/o ADQ & 4.4 & 27.65   \\
    % $\mathbf{Our (body 6)}$ & 9.3 & 27.988 & QuantNAS(4,8)  \\
    \hline
\end{tabular}

    \label{tab:quantized}
\end{table}

% TODO
% Our (body 3) quantized architecture with body block repeated 3 times is depicted in Figure~\ref{fig:arch_examples_q}.

\subsection{QuantNAS vs. NAS + fixed quantization}
We also look at whether a joint selection of architecture and bit level - \textit{mix precision setting} is better than neural architecture search for a single fixed bit level - \textit{uniform quantization setting}. 

% \ik{Parentheses are rarely used in English. It is better to reformulate the sentence, or use comma, or use dash to stress something}

We apply QuantNAS to the RFDN architecture in three distinct settings, each varying in the available bit options for each block. The first setting exclusively searches for 4-bit blocks, the second explores 8-bit blocks, and the third provides the flexibility to select either 4- or 8-bit operations for each block. 

\begin{figure}[ht]
\begin{center}
 \includegraphics[scale=0.33]{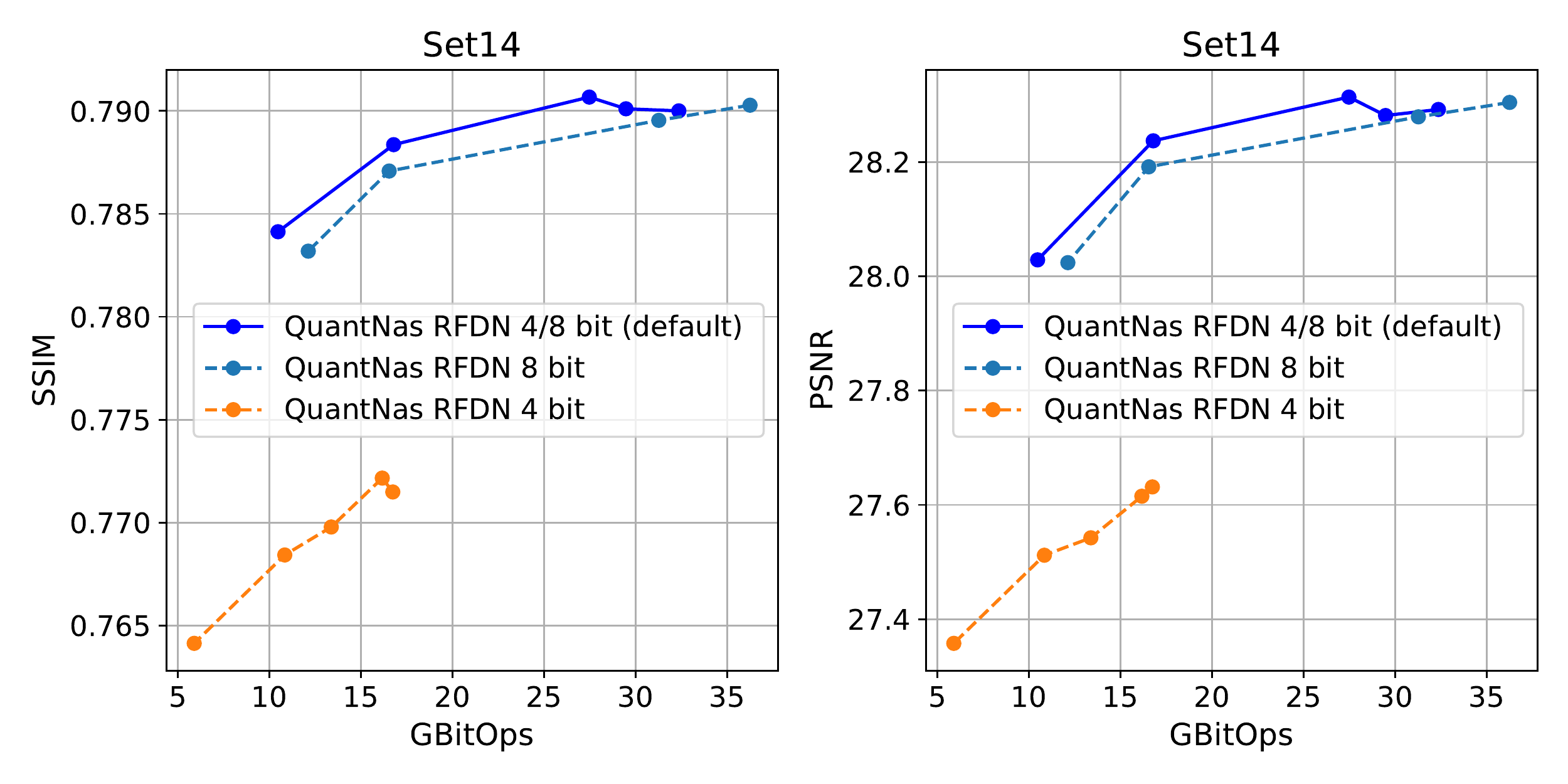}
\end{center}
\caption{
NAS + Mixed precision vs. NAS + uniform quantization. We conduct identical search for QuantNAS RFDN, but with the flexibility to search for blocks using fixed 4 bits, 8 bits, or both 4 and 8 bits simultaneously. Results are presented using the Set14 dataset via SSIM and PSNR metrics.
} 
\label{fig:rfdn_qnas_nas}
\end{figure}

Results are shown in Figure ~\ref{fig:rfdn_qnas_nas}.
The graph clearly demonstrates that broadening the search space to include mixed bit width (4/8 bits) consistently leads to the discovery of superior models.
It is worth noting that the Pareto plots for various metrics, such as SSIM and PSNR, exhibit remarkably similar results. This trend persists across all experiments.

\begin{figure}[ht]
\begin{center}
 {\includegraphics[scale=0.30]{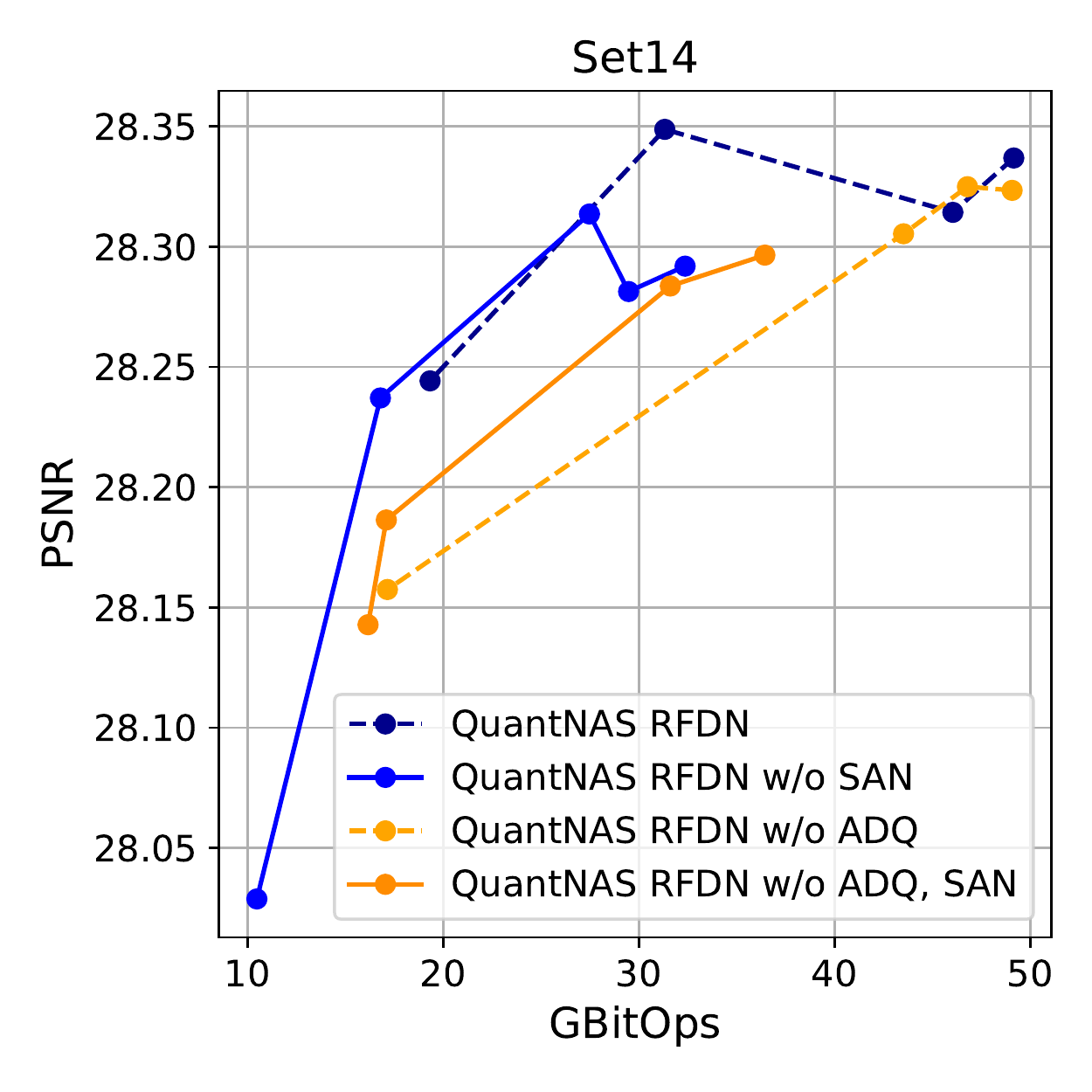}
 \includegraphics[scale=0.3]{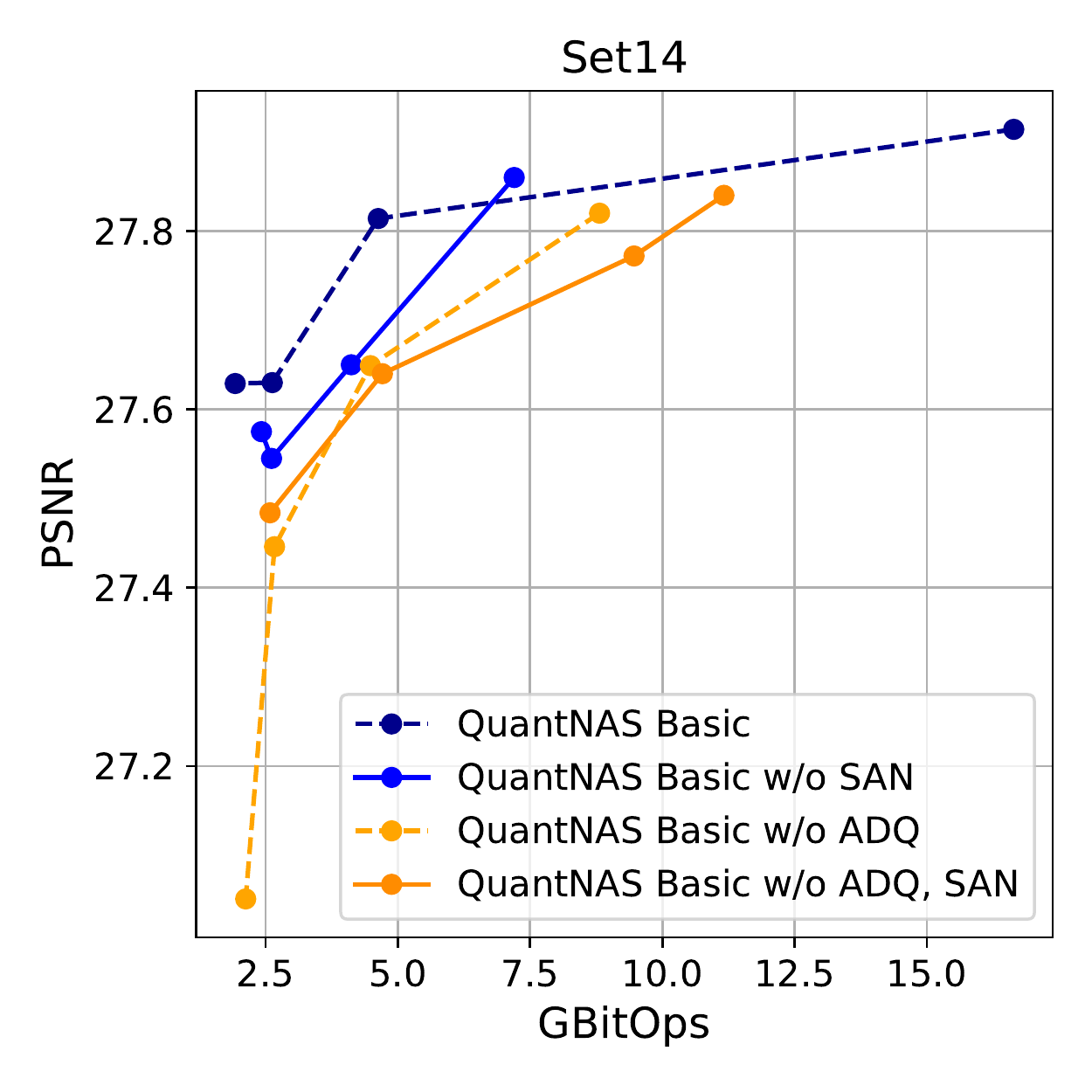}}%
\end{center}
\caption{Comparison of different NAS options: vanilla, without SAN, without ADQ, and without SAN and ADQ settings. Without SAN means that we use quantization with shared weights. Metrics are for the Set14 dataset. Left - QuantNAS RFDN, right - QuantNas Basic search space.}
% \ik{1) Labels seem to be confusing. What does QuantNAS exactly mean? From Figure 5-6, I thought that QuantNAS already includes SAN and AdaDM, but this figure contradicts it. Please, clearly state in the beginning of this section what is QuantNAS and change labels in this plot. 2) Note that somewhere you write QuantNas and somewhere QuantNAS.}
\label{fig:ablation}
\end{figure}

\subsection{Adaptive Deviation for Quantization}
\label{sec:adm}
%One of the possible reasons for it is that BN stabilizes search procedure and AdaDM mitigates problems caused by BN.
We start with comparing the effect of AdaDM \cite{AdaDM} and ADQ on three architectures randomly sampled from our search space.
Table~\ref{tab:adadm} shows that both original AdaDM and Batch Normalization hurt the final performance, while ADQ improves PSNR scores. 

\begin{table}
\caption{PSNR of SR models with scaling factor 4 for Set14 dataset. M1 and M2 are two arbitrary mixed precision models randomly sampled from our search space.}
\begin{center}
\begin{tabular}{llll}   
    %\emph{some text}     \\ \midrule
    \hline 
     Model & Model M1 & Model M2 \\
    \hline
    Without Batch Norm  &  27.55 & 28.00  \\
    With Batch Norm & 27.00 & 27.16 \\ 
    Original AdaDM & 27.33 & 27.84 \\
    Our AdaDM & $\mathbf{27.68}$ & $\mathbf{28.05}$ \\
    \hline
\end{tabular}
\end{center}
    \label{tab:adadm}
\end{table}

In Figure~\ref{fig:ablation}, we can see that architectures found with ADQ perform better in terms of both PSNR and BitOPs, highlighting the clear advantage of using ADQ in the search procedure for both our custom search space and RFDNs.

\subsection{Search Against Noise}
\paragraph{Quality}

The results shown in Figure~\ref{fig:ablation} also demonstrate the contribution of SAN to our method. 

Provided metrics demonstrate that SAN serves as a reasonable and effective replacement for direct quantization. Furthermore, in the setting involving our custom search space, SAN consistently enhances the search procedure, and when combined with ADQ, it yields a distinct improvement for RFDN.

% \ik{I tried SAN with EdMIPS and it worked well for me. But due to stability, apparently, it is possible to combine mixed-precision search and NAS. To me, using SAN in bit width searching is still the most significant contribution. But this is not clearly stated in introduction, discussion, and results are little. It would be nice to further study it and give more evidences.}

\begin{figure}[ht]
\begin{center}
 \includegraphics[scale=0.35]{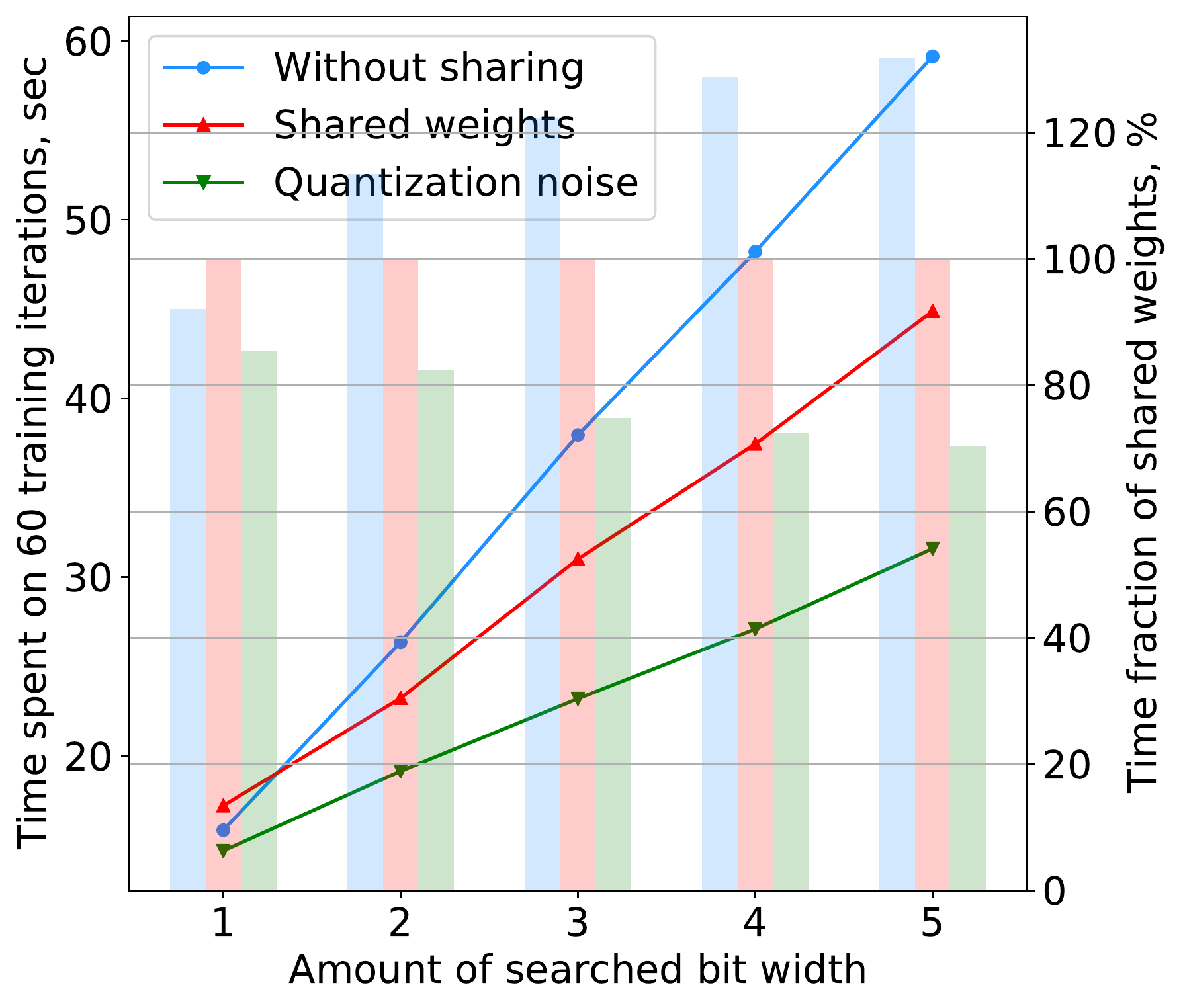}
\end{center}
\caption{Time comparison of quantization noise and weights sharing strategy during the search phase of quantization-aware NAS. Y-axis (on the left) shows time spent on 60 training iterations (line plot). The secondary Y-axis (on the right) presents the time fraction of SW strategy (bar plot). 
} 
\label{fig:timeeffectiveness}
\end{figure}

\paragraph{Time efficiency}
To demonstrate the time efficiency of our approach, we measured the average training time for three quantization methods: without weight sharing, with weight sharing used by EdMIPS, and employing search against quantization noise used by QuantNAS with SAN.

SAN reduces computation during the search phase, avoiding the need for quantizing each bit level individually. 
We ran the same experiment with varying numbers of searched quantization bits. For uniform quantization, the number of searched bit widths is 1, while for mixed precision (4 or 8 bits for each block) it is 2.

Figure~\ref{fig:timeeffectiveness} shows the advantage of SAN in training time. As the number of searched bits grows, so does the advantage. On average, we get up to $30\%$ speedup.

\subsection{Entropy regularization} 
\label{sec:entropy}

In this section, we provide evidence that the entropy regularization helps a NAS procedure and give details on the source of these improvements.

We consider three settings to compare QuantNas with and without Entropy regularization:
(A) small search space, SGD optimizer; (B) big search space, Adam~\cite{Adam} optimizer; and (C) small search space, Adam~\cite{Adam} optimizer. All the experiments were performed for full precision search. For small and big search spaces, we refer to Appendix \ref{sec: search_spaces}.
We perform the search without hardware penalty to analyze the effect of the entropy penalty. 

Quantitative results for Entropy regularization are in Table \ref{tab:entropy}. Entropy regularization improves performance in terms of PSNR for all the experiments.

\begin{figure*}[ht]
\center{\includegraphics[scale=0.35]{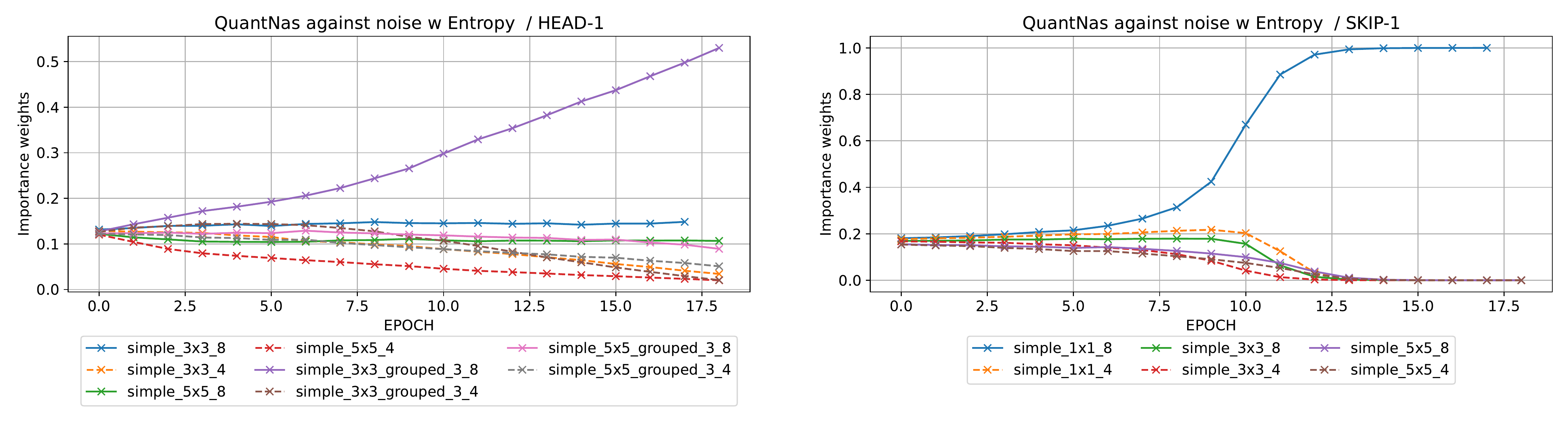}}
\center{\includegraphics[scale=0.35]{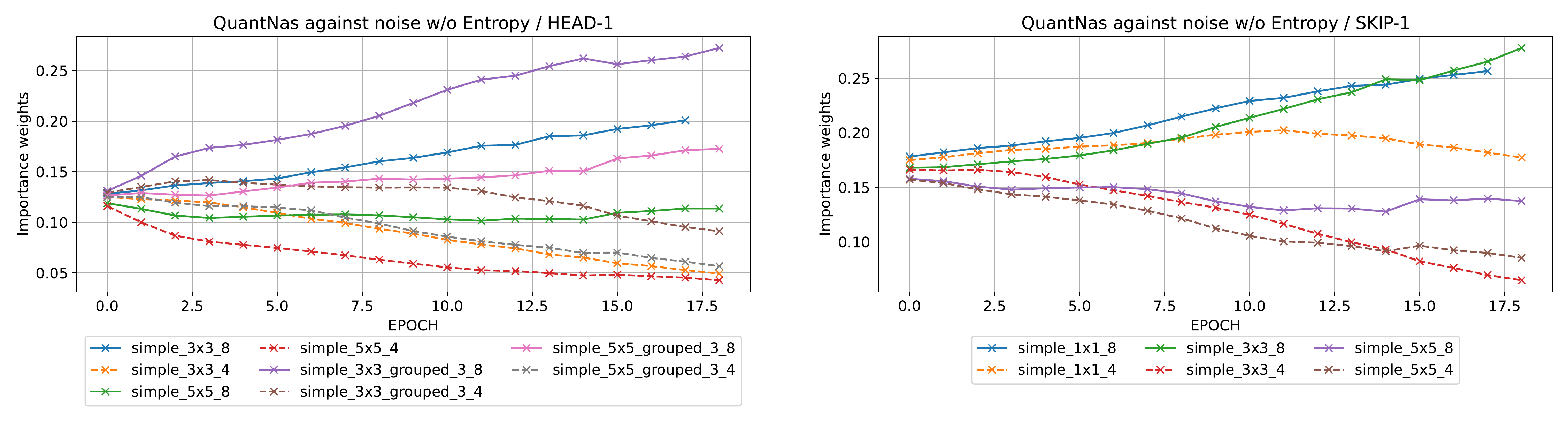}}
\caption{Dynamics of importance weights for different operations through epochs for QuantNAS. For 8 and 4 bits, we use solid and dashed lines, respectively. Usage of entropy sparsification (top) allows for selecting a single most relevant block with high importance c.t. variants without entropy sparsification (bottom).} 
% Supernet sparsification with entropy regularization on the top for two layers HEAD-1 and a parallel convolutional  layer in BODY (SKIP-1), regularization value is set to 1e-3 (on the top). 
\label{fig:alphas}
\end{figure*}

Figure~\ref{fig:alphas} demonstrates dynamics of operations importance for joint NAS with quantization for 4 and 8 bits. 4 bits edges are depicted in dashed lines. Only two layers are depicted: the first layer for the head (HEAD) block and the skip (SKIP) layer for the body block. With entropy regularization, the most important block is evident from its important weight value($\alpha$ from ~\eqref{eq:super_net} ). 
Without entropy regularization, we have no clear most important block.
So, our search procedure has two properties: (a) the input to the following layer is mostly produced as the output of a single operation from the previous layer; (b) an architecture at final search epochs is very close to the architecture obtained after selecting only one operation per layer with the highest importance value.
% \ik{I do not know if we need to add it, but I guess entropy regularization helps because of EdMIPS. In my experience (see that One-Shot paper), when sampling is used (aka DNAS), entropy should enter loss with an opposite sign and a small value. This is because sampling in the end selects the best models, and larger entrpy helps to keep all options on the table.}

\begin{table}
\centering
\caption{PSNR/GFLOPs values of search procedure with and without Entropy regularization. Models were searched in different settings A, B, and C. 
% \ik{To me, it looks strange that with Entropy B finds a much smaller models, but quality is best. Maybe some clarification in text or appendix is needed.}
}
\begin{tabular}{llll}   
    %\emph{some text}     \\ \midrule
    \hline
    Training settings & without Entropy & with Entropy  &\\
        \hline
    A   & 27.99 / 111 &  $\mathbf{28.10}$ / 206 \\
    B   & 28.00 / 30 &  $\mathbf{28.12}$ / 19  \\
    C   & 27.92 / 61 & $\mathbf{28.11}$ / 321\\
%        \hline
    %& PSNR / BitOPs & \\
    %D   & 27.84 / 2128  & $\mathbf{27.93}$ / 2272 \\
    \hline
\end{tabular}
\label{tab:entropy}
\end{table}

\section{Discussion}

% \ik{I like the passage in abstract that "the proposed approximations are better for search procedure than direct model quantization". Please elaborate on why do you think it achieves better results? My guess is because it uses the correct derivative instead of STE which helps NAS stability. Can we actually highlight it in results also?}

% \ik{Also, the reasons and hypothesis of some unexpected results can be discussed in this section. For example, why do we see in Figure 7 (left) the drop in PSNR for "QuantNas RFDN AdaDM+SAN and especially for "QuantNas RFDN AdaDM". Could this be the evidence of stability due to SAN?}

% \ik{Finally, it would be insightful to analyze found architectures. What is the relationship between operations and bit widths? Do we confirm OQAT result?}

We demonstrate that with  SAN, we are able to achieve a close approximation of direct quantization. Additionally, SAN produces superior results, potentially attributed to its differentiable reparametrization. However, the stochastic nature introduced by randomly sampled quantization noise makes the SAN procedure less stable. Interestingly, our findings reveal that when combined with ADQ, SAN consistently delivers improved outcomes, whereas using SAN alone may result in suboptimal solutions. In the subsequent section (Section~\ref{sec:architechure_analysis}), we conduct a thorough analysis of the  architectures and delve into further insights.

% \ik{I think this section should be renamed to Discussion. Limitations can be mentioned in the end of it. It is very important to disscuss all the results. For example, given Table I, I wonder why our search space is worse then RFDN search space? I will think more about the questions we could answer here.}

We have successfully showcased the efficacy of our procedure in two search spaces, indicating its potential applicability to other search spaces as well.  \textit{RFDN search space} consistently outperforms our \textit{Basic search space} due to the incorporation of various technical solutions, including Residual Feature Distillation. It is worth noting that the development of such search spaces requires considerable effort. However, our results demonstrate that it is possible to design a customized search space based on an existing architecture, resulting in improved quality and efficiency.

For our QuantNAS procedure, the overall NAS limitation applies: the computational demand for joint optimization of many architectures is high. The search procedure takes about $24$ hours to finish for a single GPU TITAN RTX. Moreover, obtaining the full Pareto frontier requires running the same experiment multiple times.  

In Figure~\ref{fig:ablation}, all most right points (within one experiment/color) have $0$ hardware penalty. It clearly shows that limited search space creates an upper bound for the top model performance. Therefore, results for our search space do not fall within the same BitOps range as SRResNet. 

We found that our procedure is sensitive to hyperparameters. In particular, optimal coefficients for hardware penalty and entropy regularization can vary across different search settings. 
Moreover, we expect that there is a connection between optimal coefficients for the hardware penalty, entropy regularization, and search space size.
Different strategies or search settings require different values of hardware penalties. 
Applying the same set of values for different settings might not be the best option, but it is not straightforward as how to determine them beforehand.

\section{Conclusion}
We introduce a novel method called QuantNAS, which combines NAS and mixed-precision quantization to obtain highly efficient and accurate architectures for Super-Resolution (SR) tasks. To the best of our knowledge, we are the first to extensively explore the integration of NAS with mixed-precision search for SR.
 
We propose the following techniques to enhance our search procedure: (1) The entropy regularization to avoid co-adaptation in supernets during differentiable search; (2) differentiable SAN procedure; and (3) ADQ module which helps to alleviate problems caused by Batch Norm blocks in super-resolution models.

We demonstrate the versatility of our method by applying it to various search spaces. In particular, we conduct experiments using search space based on the computationally efficient SR model RFDN.

Our experiments clearly indicate that the joint NAS and mixed-precision quantization procedure outperforms using NAS or mixed-precision quantization alone.

Furthermore, when compared to the mixed-precision quantization of popular SR architectures with EdMIPS \cite{edmips}, our search consistently yields better solutions. Additionally, SAN search approach is up to 30\% faster than a shared-weights approach.

% In summary, our method showcases impressive performance on standard SR tasks, surpassing competing methods in terms of both quality and speed.

% \ik{References are not polished, please make it look in a single style. E.g., somewhere "nas" is written in small letters, somewhere you write "Proceedings of the IEEE Conference on Computer Vision and Pattern Recognition", somewhere "Proceedings of the IEEE conference on computer vision and pattern recognition", somewhere "Proceedings of the IEEE conference on computer vision and pattern recognition", etc. Also check "arXiv preprint arXiv:1902.08153" and in other place just "arXiv:2104.09987v2". All of it should be in the same style before submission}

% \section{Acknowledgement}
% Alexey Zaytsev and 
% Evgeny Burnaev were supported by the Russian Foundation for Basic Research grant 21-51-12005 NNIO\_a.
\clearpage
\bibliographystyle{IEEEtran}
\bibliography{source}

% {\small
% \bibliographystyle{ieee_fullname}
% \bibliography{egbib}
% }

% \clearpage
% \subsection{References}
% \bibliographystyle{plain}
% \bibliography{references}

\clearpage

%\section*{Appendix}
\appendix

\section*{Appendix}
\section{Technical details}
%\label{sec:techdetails}
During the search phase, we consider architectures with a fixed number of  channels for each operation unless channel size is changed due to operations properties. For \textit{Basic search space}, number of channels is set to 36, and for \textit{RFDN search space} number of channels is set to 48.
The search is performed for 20 epochs. To update the weights of the supernet, we utilize the following hyperparameters: batch size of 16, an initial learning rate (lr) of 1e-3, a cosine learning rate scheduler, SGD with a momentum of 0.9, and a weight decay of 3e-7. When updating the alphas, we employ a fixed lr of 3e-4 and no weight decay.

During the training phase, an obtained architecture is optimized for 30 epochs with the following hyperparameters: batch size 16, initial lr 1e-3, and lr scheduler with the weight decay of 3e-7.

\begin{table}[b!]
\centering
\caption{Quantitative results of PSNR-oriented models with SR scaling factor 4 for Set14 dataset. $*$ results are from paper \cite{Trilevel}}
\begin{tabular}{llll}   
    %\emph{some text}     \\ \midrule
    \hline
     Method & GFLOPs  & PSNR & Search cost \\ 
    \hline
    SRResNet  & 166.0 & 28.49  & Manual   \\ 
    RFDN  & 27.14 & 28.37  & Manual   \\  
    AGD$\phantom{}^{*}$  & 140.0 & 28.40  & 1.8 GPU days   \\ 
    Trilevel NAS$\phantom{}^{*}$ & 33.3 & 28.26   & 6 GPU days   \\
    \hline
    QuantNAS Our SP & $\mathbf{29.3}$  & 28.22 & 1.2 GPU days  \\
    QuantNAS RFDN & $\mathbf{23.4}$  & 28.3 & 1.6 GPU days  \\
    
    \hline
\end{tabular}
    
    \label{tab:fullprecision}
\end{table}
\section{QuantNAS algorithm}
\begin{algorithm}
	\caption{QuantNAS steps} 
        \label{alg:quant_nas}
	\begin{algorithmic}[1]
	\STATE Initialize parameters $W$ and edge values $\alpha$
		\FOR {$iteration=1,2,\ldots, N$}
			\STATE Add QN to $W$ as in  \ref{eq:noise} and \ref{eq:super_net_quant_noise} 
			\STATE Compute the loss function $L(\alpha)$ as in \ref{sec:alpha_loss}
			\STATE  Run backpropagation to get derivatives for $\alpha$ 
			\STATE  Update $\alpha$
			\STATE  Add QN to $W$ as in  \ref{eq:noise} and \ref{eq:super_net_quant_noise} 
			\STATE Compute the loss function $L_1(W)$ 
			\STATE Run backpropagation to get derivatives for $W$
			\STATE Update $W$
		\ENDFOR
		\STATE Select edges with the highest $\alpha$
		\STATE  Train the final architecture from scratch
	\end{algorithmic} 
\end{algorithm}

\section{Comparison with existing full precision NAS for SR}
\label{sec:fpnas}

Here we examine the quality of our procedure for full precision NAS on \textit{Basic search space} without ADQ and SAN. 
 
The results are in Table \ref{tab:fullprecision}. Our search procedure achieves comparable results with TrilevelNAS\cite{Trilevel} with a relatively simpler search design and about 5 times faster search time. 
The best performing full precision architecture was found with a hardware penalty of value $1e-3$. 
This architecture is depicted in Appendix Figure~\ref{fig:arch_examples_fp}. 

Additionally, we compare results with a popular SR architecture SRResNet \cite{SRRESNET}. 
Visual examples of the obtained super-resolution pictures are presented in  Figure~\ref{fig:images} for Set14~\cite{Set14}, Set5~\cite{Set14}, Urban100~\cite{Urban100}, and Manga109~\cite{fujimoto2016manga109}  with scale factor 4.

\section{Analysis of found architectures}
\label{sec:architechure_analysis}
\begin{figure*}[h]
\center{\includegraphics[scale=0.148]{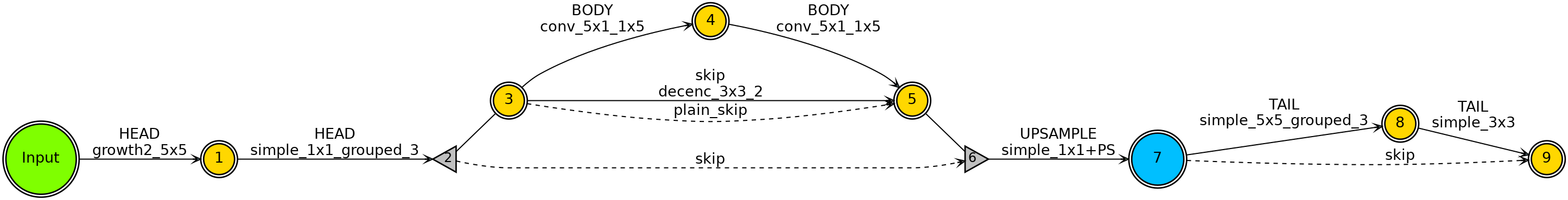}}
\caption{Our best FP (full precision) architecture, 29.3 GFLOPs (image size 265x265), PSNR: 28.22 dB.  PSNR was computed on Set14. Body block is repeated three times for both architectures. }
\label{fig:arch_examples_fp}
\end{figure*}

\begin{figure*}[ht]
\center{\includegraphics[scale=0.17]{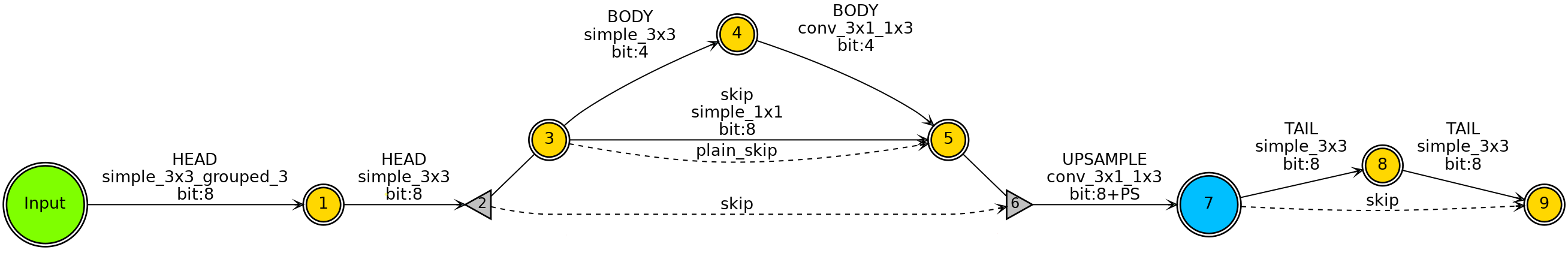}}
\center{\includegraphics[scale=0.14]{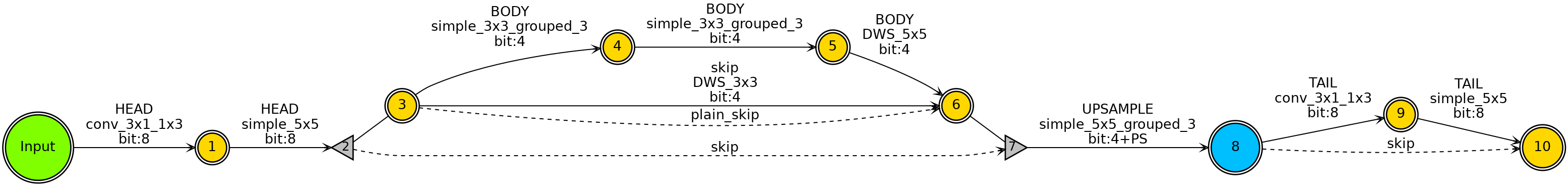}}
\center{\includegraphics[scale=0.15]{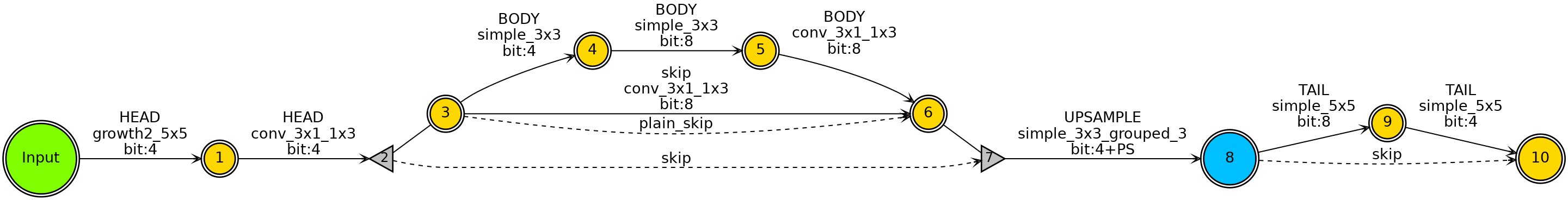}}
\caption{Examples of quantized architechtures. PSNR from top to bottom: 27.814 dB, 27.2 db, 24.8 db. On the top is our quantized architecture (body 3),  more details are given in Table~\ref{tab:quantized}. PSNR was computed on Set14 with scale 4. Body block is repeated three times for all the architectures.  Architecture on the bottom was sampled randomly.}
\label{fig:arch_examples_q}
\end{figure*}

We conducted an analysis of architectures discovered within our \textit{Basic search space}, and exemplary architectures are presented in Figures~\ref{fig:arch_examples_fp}~and~\ref{fig:arch_examples_q} for full precision and quantized models, respectively. Our observations indicate that architectures with higher performance tend to have higher bit values for the first and last layers. Notably, the quantization of the first layer has a significant impact on model performance, as it results in substantial information loss due to the quantization of incoming signal. Additionally, we found that intermediate body blocks typically exhibit lower bit values.

\section{Random search}
\label{sec:random_search}
\begin{figure}[ht]
 {\includegraphics[scale=0.40]{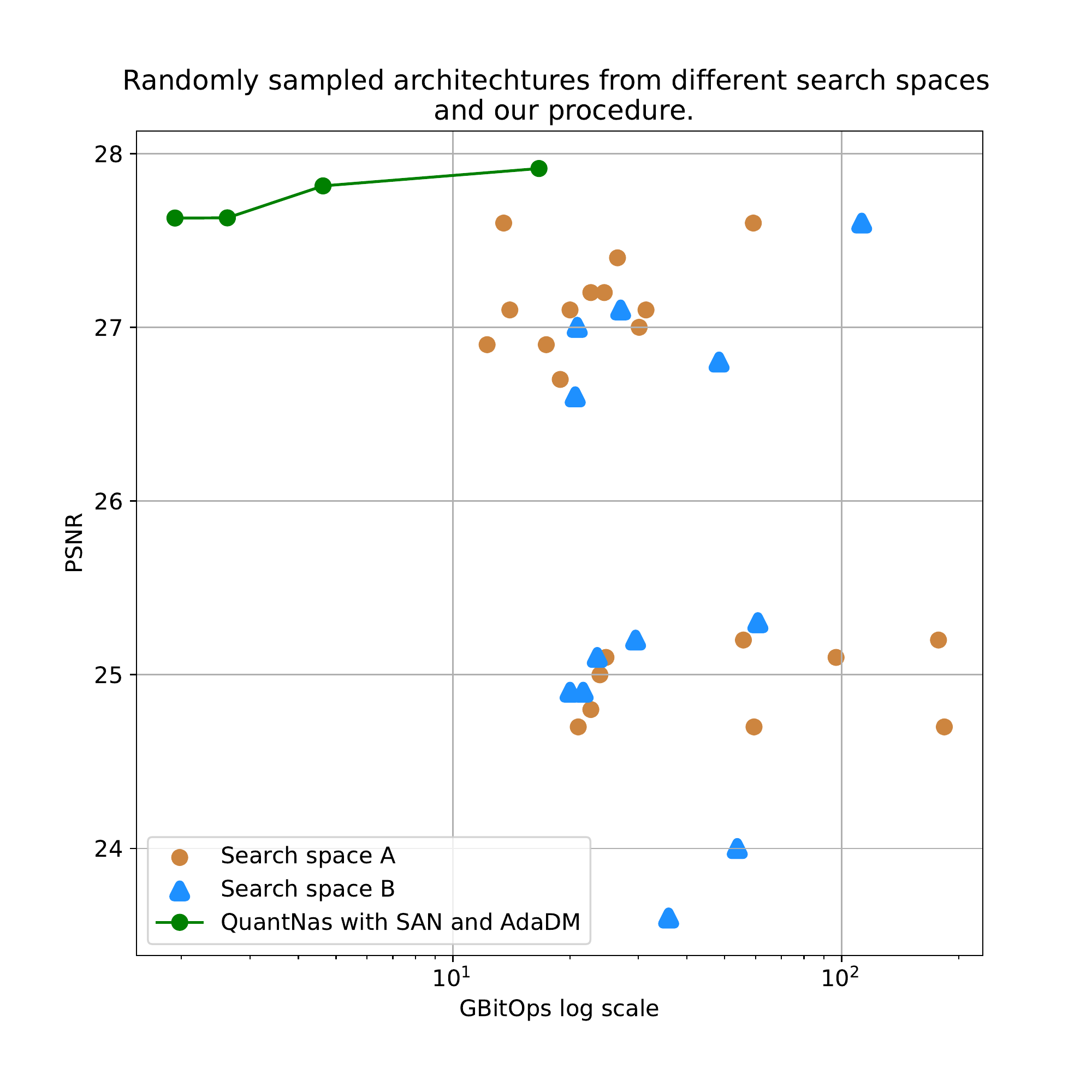}}
\caption{Randomly sampled architectures from two search spaces. The search spaces are described in the corresponding section. PSNR was computed on Set14 and BitOPs for image size 256x256. We observe that two search spaces provide slightly different results with random sampling. Results in green for architecture search were obtained with Big search space - A. Two clusters above and below 26 PSNR line attribute to 8- and 4-bit quantization of the first layer.} 
\label{fig:random_sampling}
\end{figure}

In Figure~\ref{fig:random_sampling}, we conducted a comparison between our procedure and randomly sampled architectures on the \textit{Basic search space}. The results indicate that our procedure significantly outperforms random search. Notably, there are two distinct clusters above and below 26 PSNR line, which correspond to models with 8- and 4-bit quantization of the first layers.

\section{Entropy schedule}
\label{sec:app_entropy}
\begin{figure}[ht]
\center{\includegraphics[scale=0.6]{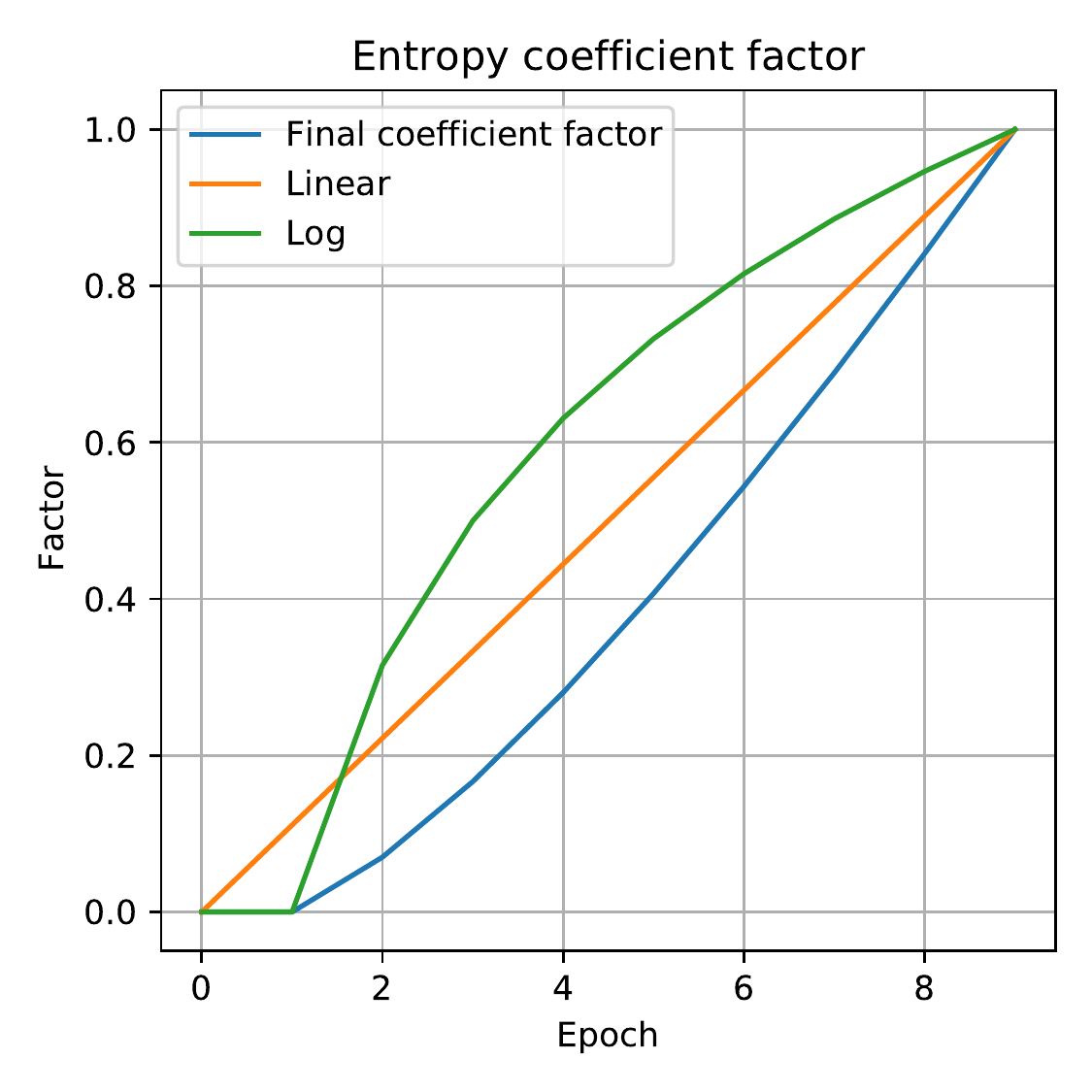}}
\caption{Entropy coefficient regularization is a product of log and linear functions.}
\label{fig:ent_coef}
\end{figure}

For entropy regularization, we gradually increase the regularization value $\alpha$ according to Figure \ref{fig:ent_coef}, and for the first two epochs, regularization is zero. Entropy regularization is multiplied by an initial coefficient and coefficient factor. Initial coefficients are 1e-3 and 1e-4 for experiments with full precision and the quantization-aware search.
% TODO: add a comparison of linear and log options

% \section{Dynamic of $\alpha$ values}

% In QuantNAS, an operation can have a different bit width value.
% Thus, with quantization, we enrich our NAS search space. 
% Additional $\alpha$ values correspond to different bits of an operation.

% The dynamics of $\alpha$ values for different NAS is in Figure \ref{fig:alphas}.
% Head and tail blocks follow similar behaviour.
% With entropy regularization turned on, we start with close values of $\alpha$, but the end of the optimization we get a single leader with order of magnitude higher $\alpha$ value.
% Without entropy $\alpha$ values remain close until the end of optimization, that lead under-training of each block and higher overall training cost with possible performance degradation due to joint training of diverse components.

% Figure~\ref{fig:alphas} shows difference in training dynamics with and without sparsity regularization. Entropy regularization helps to determine one single function. Also, it is interesting to note that sometimes identical blocks with different bit values have similar importance. It can be seen in Figure \ref{fig:alphas} second from the top experiment for body and skip blocks.

 \begin{figure}[ht]
 \center{\includegraphics[scale=0.6]{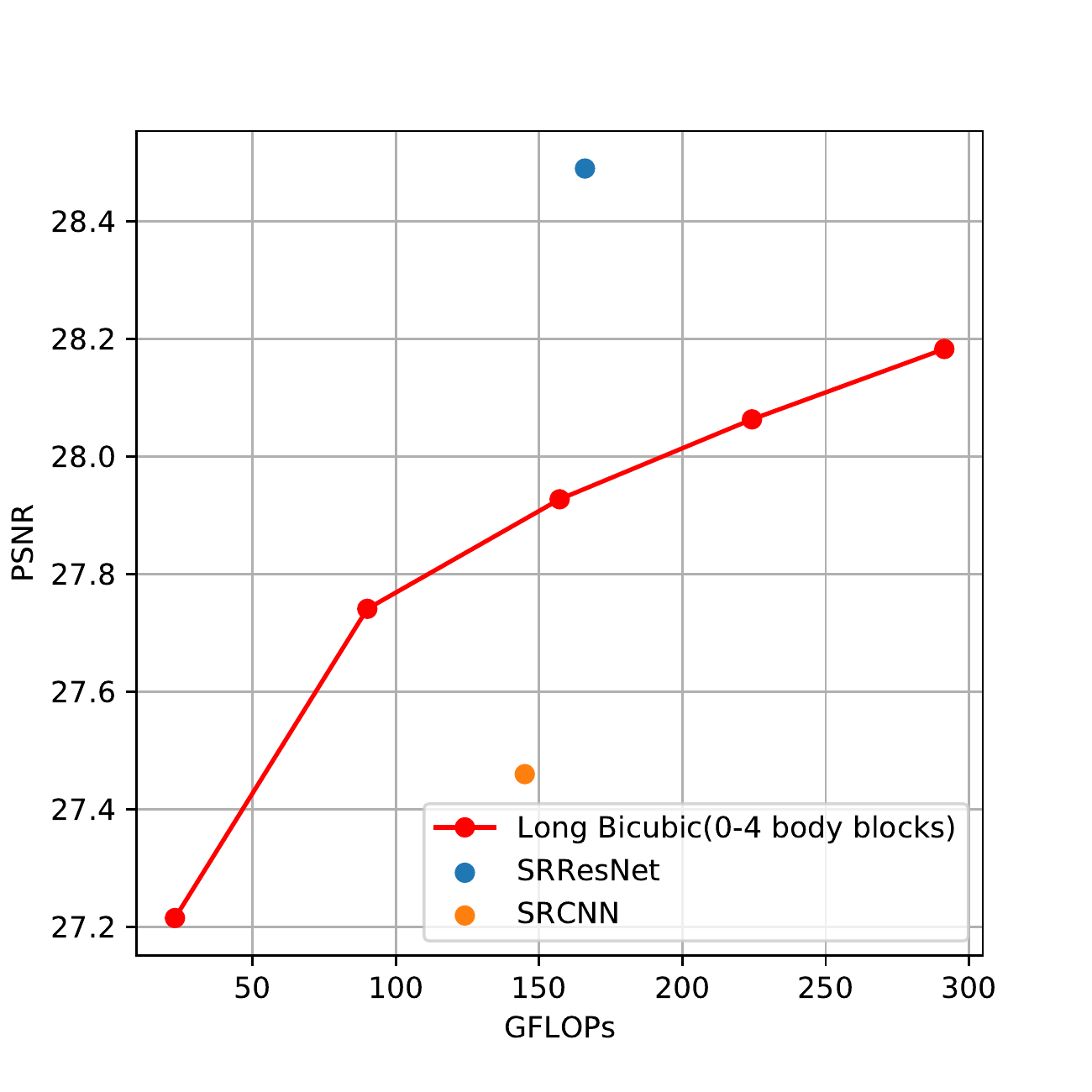}}
 \caption{Orange point is the original SRCNN \cite{SRCNN}, and blue point is SRResNet \cite{SRRESNET}. For Long Bicubic, we initially upscale an image with bicubic interpolation and then add an efficient block found in our experiments. The block consists of 3 convolutions layers with 32 filters and is added 1, 2, 3, 4 times. PSNR is reported on Set14.}
 \label{fig:bicubic}
 \end{figure}
\section{Scaling SR models with initial up-sampling}
To maintain good computational efficiency, it is common for SR models to operate on down-sampled images and then up-sample them with some up-sampling layers. This idea was introduced first in ESPCN~\cite{ESPCN}. Since then, there were not many works in the literature exploring SR models on initially up-scaled images. 

% On another hand, we observed that having a few convolutional layers after up-sampling operation significantly benefits model performance. It may suggest that these two approaches may have different properties.  In our search space, demonstrated in Figure~\ref{fig:search_design}, we have two layers of convolutions called Tail placed right after up-sampling block.  

Therefore,  we were interested in how this approach scales in terms of quality and computational efficiency given arbitrary many layers. Results are presented in Figure~\ref{fig:bicubic}. We start with one fixed block, similar to our body block in Figure~\ref{fig:search_design}, and then increase it by one each time. We compare our results with SRResNet~\cite{SRRESNET} and SRCNN~\cite{SRCNN}. As we can see, SRResNet~\cite{SRRESNET} operates on down-scaled images and yields better results given the same computational complexity. 

% However, we believe that this direction can be studied further.

\begin{figure}[ht]
   \begin{center}
      {\includegraphics[scale=0.4]{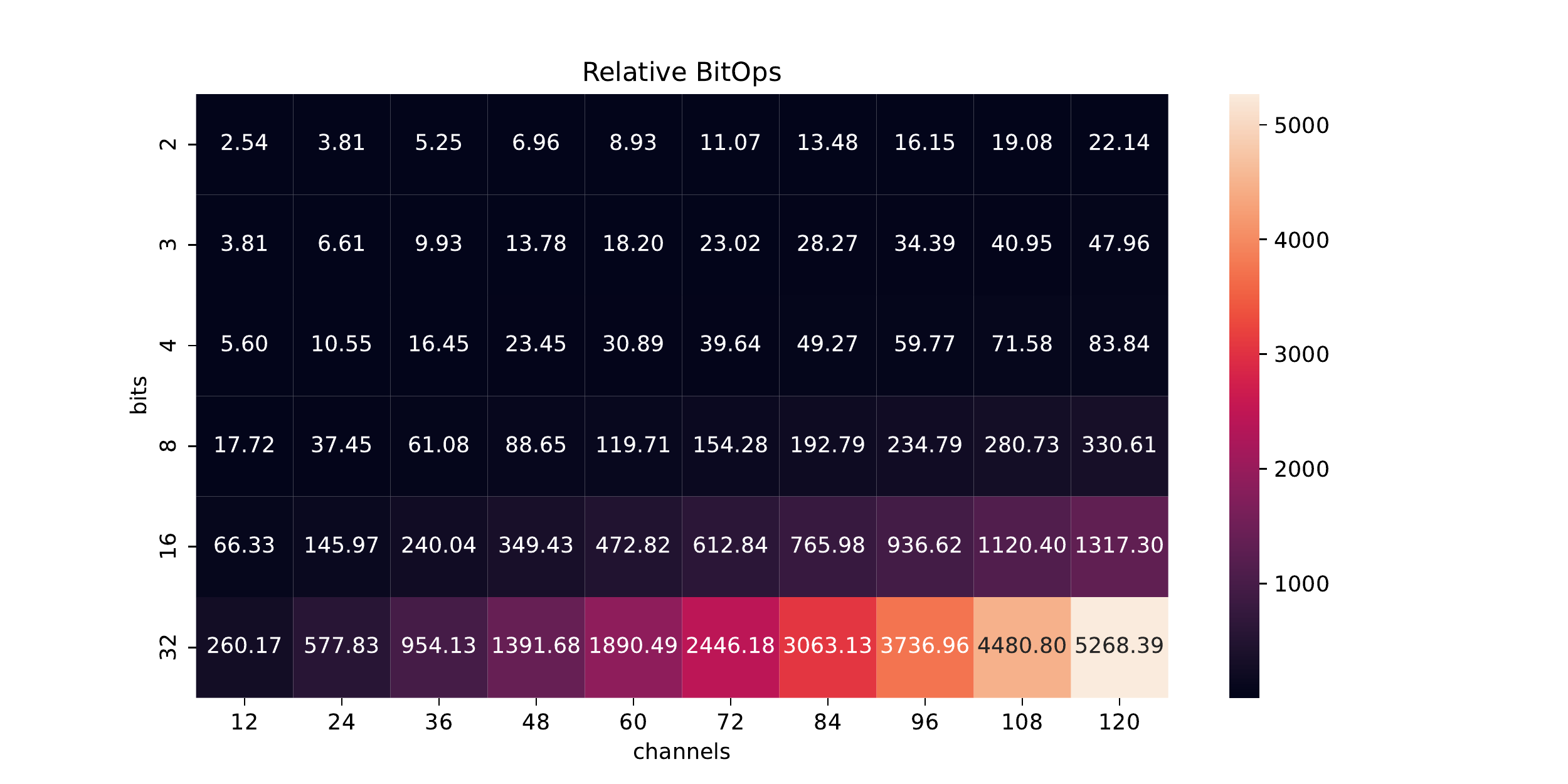}}
      {\includegraphics[scale=0.4]{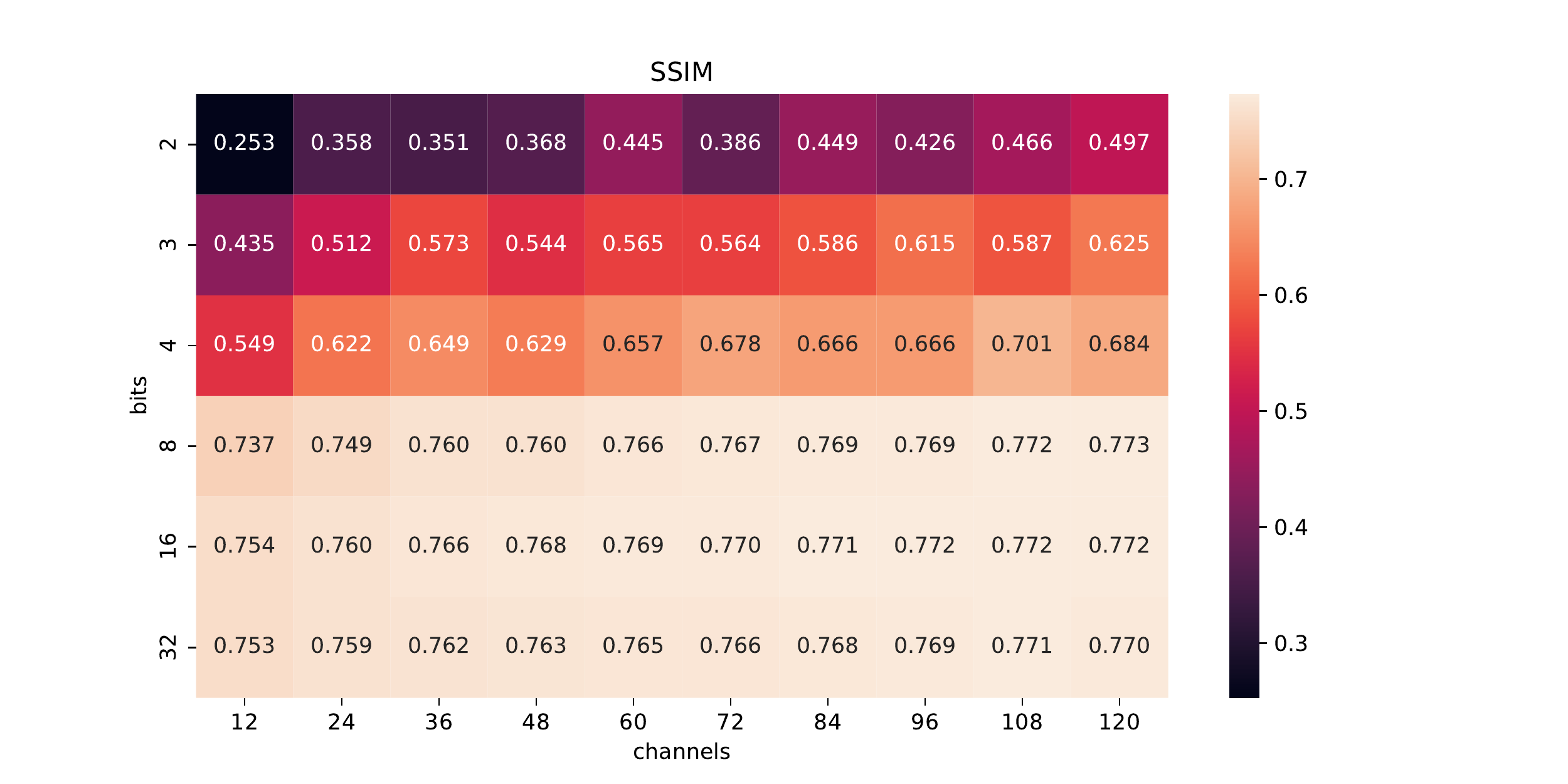}}
    \end{center}
\caption{Performance comparison with  different bit values and number of channels on the ESPCN model. All layers are uniformly quantized, except for the first layer, which is fixed with with 32 bits. BitOps values are scaled and relative values are reported.}
\label{fig:espcn_grid_bits}
\end{figure}

\section{Determining the Importance of Bits and Channels}
In Figure~\ref{fig:espcn_grid_bits}, we conducted an analysis to determine the relative importance of bits and channels in model performance. Our findings are as follows. The results reveal that using 8-bit quantization yields comparable performance to that of 16- and 32-bit quantization, while providing marginal computational efficiency gains. This suggests that 8-bit quantization is a viable option for achieving efficient performance. Additionally, we observed that increasing the number of channels in a model comes at a higher cost and is not practical. As a result, it is essential to explore alternative optimization approaches to enhance model performance, rather than rely solely on channel scaling. One such approaches is feature distillation used in RFDN. 

Considering these findings, we decided not to include the number of channels in our search space, since it has a less significant impact on model performance.

\section{Search space}

\subsection{Single-path search space}
\label{sec:single_path}
 There are several ways to select directed acyclic sub-graph from a supernet. DARTS \cite{Darts} uses Multi-Path strategy - one node can have several input edges.  Such a strategy makes a search space significantly larger. In our work, we use Single-Path strategy - each searchable layer in the network can choose only one operation from the layer-wise search space (Figure \ref{fig:dag_supernet}).
 %\ik{Such a strategy makes a search space significantly larger}
 It has been shown in FBNet \cite{FBNet} that simpler Single-Path approach yields are comparable with Multi-Path approach results for classification problems. Additionally, since it aligns more with SR search design in our work, we use Single-Path approach.

\label{sec: search_spaces}
We have a fixed number of channels for all the layers unless specified. 
For detailed operations description, we refer to  \href{https://anonymous.4open.science/r/QuanToaster/README.md}{our code}. 

% \begin{itemize}
%  \item \textbf{Head} : two layers, first layer computes images with three channels;
%  \item \textbf{Body}: a repeatable cell with a skip connection and three layers; 
%  \item \textbf{Skip}: a single convolutional layer within body cell connecting input and output (do not confuse with conventional skip operation);
%  \item \textbf{Upsample}: one layer before pixel shuffle, this layer increases a number of channels before pixel shuffle and pixel shuffle operation outputs 3 channels image; 
%  \item \textbf{Tail} : two layers, final layer outputs images  with 3 channels. 
% \end{itemize}

\subsection{Search space  (Big - A):}
This search space was used for full precision experiments, unless specified.
\label{sec:search_spaces}
Possible operations block-wise:
\begin{itemize}
 \item \textbf{Head} 8 operations:   simple 3x3, simple 5x5, growth2 5x5, growth2 3x3, simple 3x3 grouped 3, simple 5x5 grouped 3, simple 1x1 grouped 3, simple 1x1;
 \item \textbf{Body} 7 operations:  simple 3x3, simple 5x5, simple 3x3 grouped 3, simple 5x5 grouped 3, decenc 3x3 2, decenc 5x5 2, simple 1x1 grouped 3;
 \item \textbf{Skip} 4 operations:   decenc 3x3 2, decenc 5x5 2, simple 3x3, simple 5x5;
 \item \textbf{Upsample} 12 operations:  conv 5x1 1x5, conv 3x1 1x3, simple 3x3, simple 5x5, growth2 5x5, growth2 3x3, decenc 3x3 2, decenc 5x5 2, simple 3x3 grouped 3, simple 5x5 grouped 3, simple 1x1 grouped 3, simple 1x1;
 \item \textbf{Tail} 8 operations:  simple 3x3, simple 5x5, growth2 5x5, growth2 3x3, simple 3x3 grouped 3, simple 5x5 grouped 3, simple 1x1 grouped 3, simple 1x1;
\end{itemize}

\subsection{Search space (Small - B):}
This search space was mainly used for all Quantization experiments
\label{sec:q_space}
Possible operations block-wise:
\begin{itemize}
 \item \textbf{Head} 5 operations: simple 3x3, simple 5x5, simple 3x3 grouped 3, simple 5x5 grouped 3;
 \item  \textbf{Body} 4 operations: conv 5x1 1x5, conv 3x1 1x3, simple 3x3, simple 5x5; 
 \item \textbf{Skip} 3 operations:  simple 1x1, simple 3x3, simple 5x5;
 \item \textbf{Upsample} 4 operations:  conv 5x1 1x5, conv 3x1 1x3, simple 3x3, simple 5x5;
 \item \textbf{Tail} 3 operations:  simple 1x1, simple 3x3, simple 5x5;
\end{itemize}

Conv 5x1 1x5 and conv  3x1 1x3 are depth-wise separable convolution convolutions. For operations description, we refer to our code.

%\section{Joint Search vs two steps approach}
% move to appendix 
%We compare joint QuantNas with two steps approach. Two step approach is when architecture searched first and qauntized after. We demonstrate our results for different architectures, found with our full precision NAS and joint search in Table \ref{tab:bilevel}. We find that two steps and joint approaches yield comparable results.

%\begin{table}
%     
%\begin{tabular}{@{} *7l @{}}   
    %\emph{some text}     \\ \midrule
%    \hline
%     \emph{Method} &  $\frac{PSNR}{FP}$ & \emph{GFLOPs} & $\frac{PSNR}{Quantized}$ & \emph{GBitOps} &  \emph{bit widths}\\  
%     \hline
%     JOINT (Reg: 0)  & - & -  & 28.0    & 1690 & 4,8 \\
%     JOINT (Reg: 0.005) & - & -  & 27.7    & 123  & 4,8\\
%     JOINT (Reg: 0) & - & -  & 24.7    & 40  & 2,4\\
%     Two steps (Reg: 0.005) & 28.31 & 336.31  & 27.9    & 1370 & 4,8\\
%     Two steps (Reg: 0.005) & 28.22 & 29.3  & 27.7  & 119 & 4,8\\
%     \hline
%\end{tabular}
%\caption{Comparison of joint search versus two steps approach. Regularization is specified for quantization procedure or joint search. For two steps approach first we search for an architecture using our NAS and then quantize it using mixed precision quantization. PSNR/FP and FLOPs were computed for full precision model before quantization.}
 %\label{tab:bilevel}
%\end{table}

\section{Results on other datasets}
\begin{figure*}[h]
  \includegraphics[width=1\linewidth]{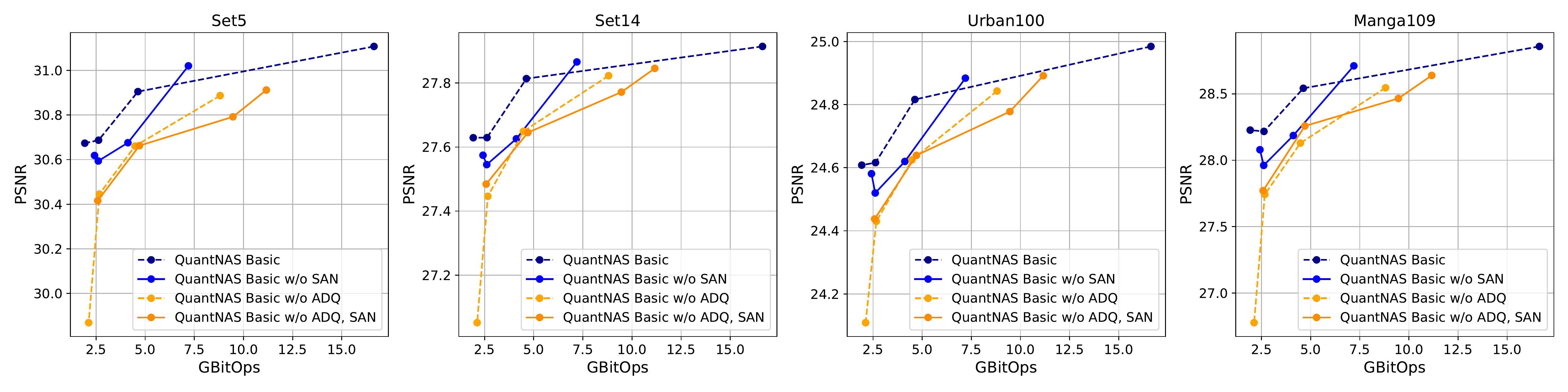}
  \includegraphics[width=1\linewidth]{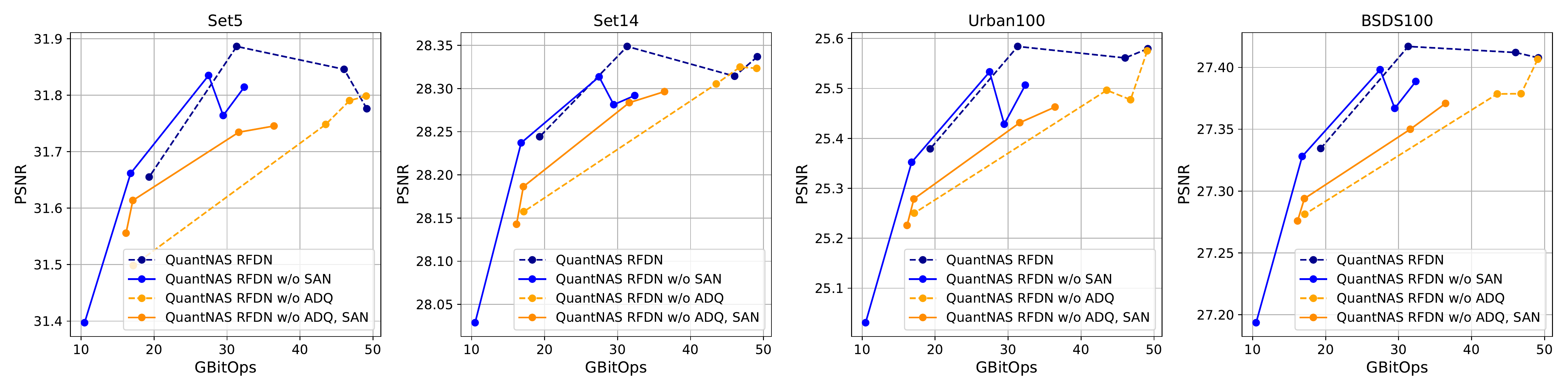}  
   \caption{The same as Figure~\ref{fig:ablation} but for different datasets.}
   \label{fig:all_datasets}
\end{figure*}

In Figure~\ref{fig:all_datasets}, we provide quantative results obtained on different test datasets: Set14~\cite{Set14}, Set5~\cite{bevilacqua2012set5}, Urban100~\cite{Urban100}, Manga109~\cite{fujimoto2016manga109}  with scale 4. 

\begin{figure*}[ht]
   
%   \fbox{\rule{0pt}{0.5in} \rule{0.9\linewidth}{0pt}}
  \includegraphics[width=1\linewidth]{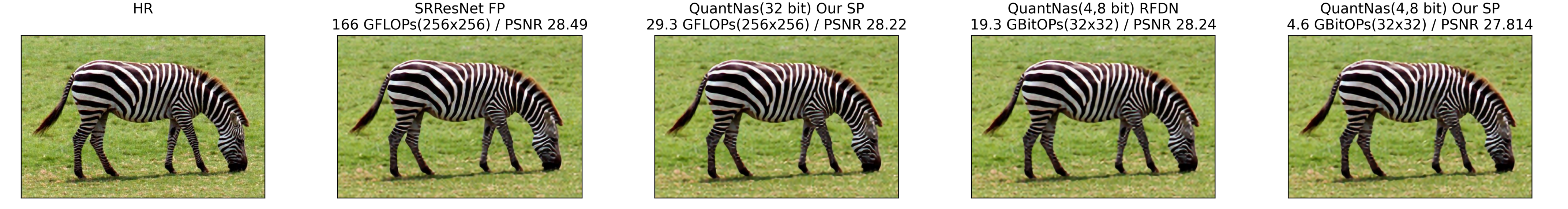}
    \includegraphics[width=1\linewidth]{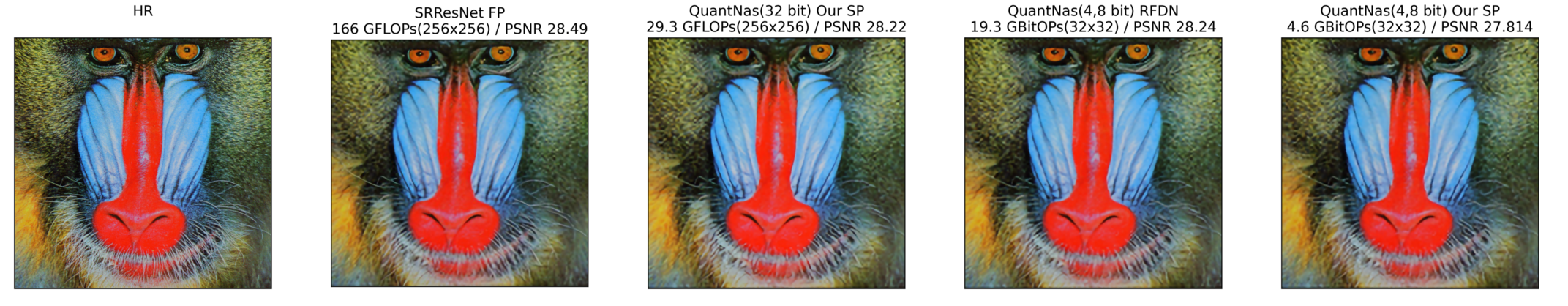}
    \includegraphics[width=1\linewidth]{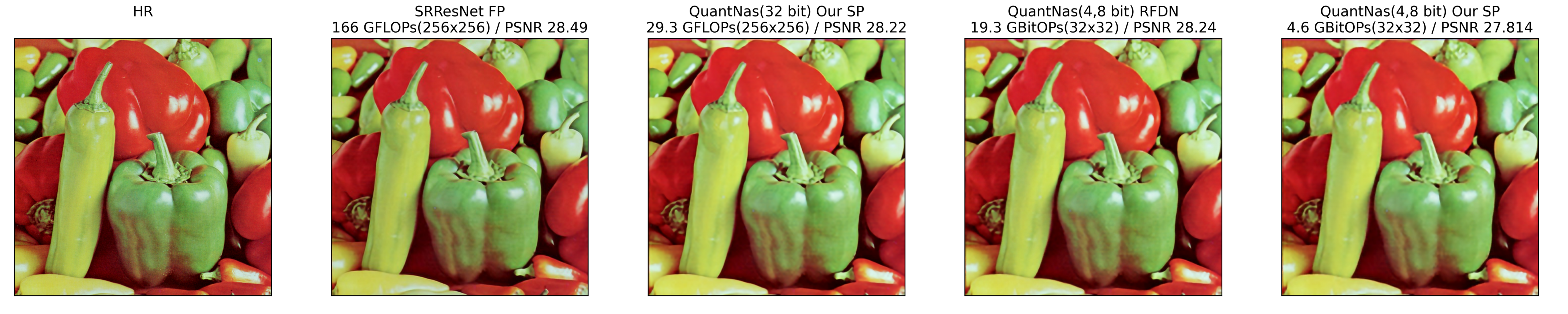}
    \includegraphics[width=1\linewidth]{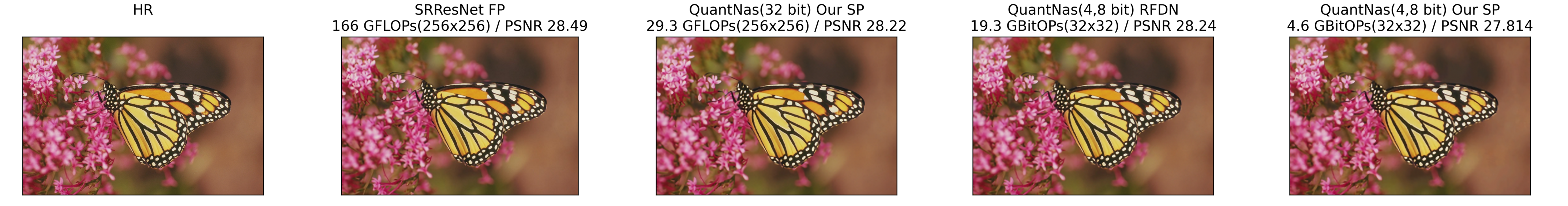}
    \includegraphics[width=1\linewidth]{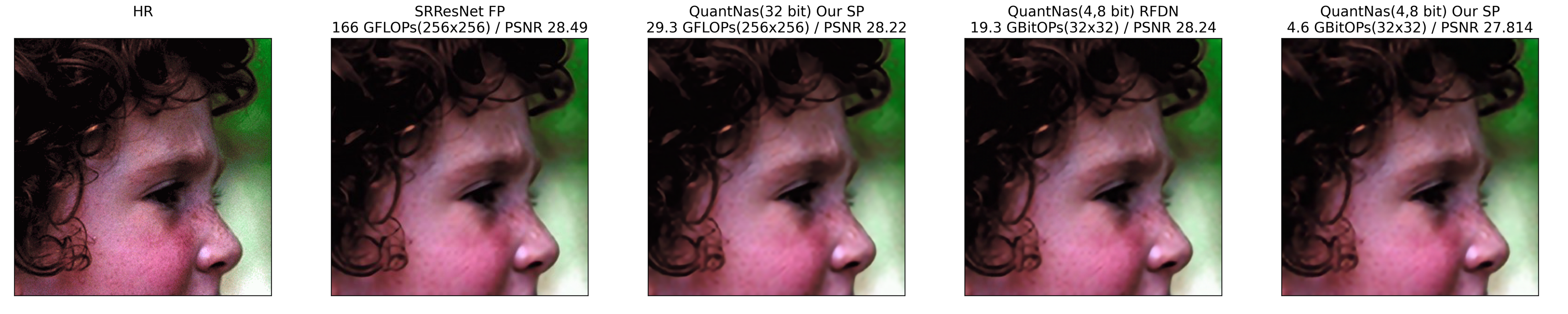}
    \includegraphics[width=1\linewidth]{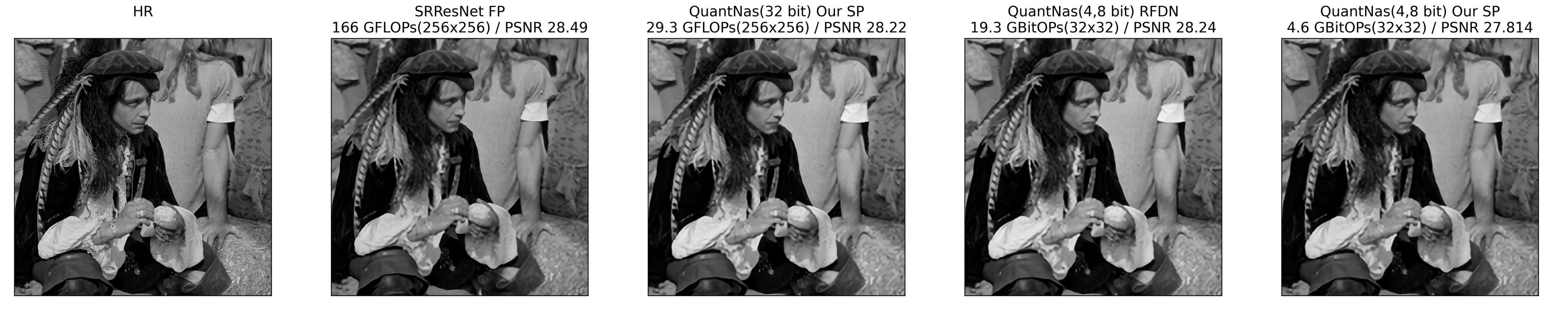}
    
  % Уместил в одну строку
%   \caption{Visual comparison of results for 1. SRResNet; 2. found full precision model; 3. found quantized model. PSNR is reported for Set14. Better view in zoom.}
\caption{Visual comparison of results for Set14. Better view in zoom. Note: we present results for quantized models with the body block repeated 3 times. Model with the body block repeated 6 times has better PSNR values (see in Table~\ref{tab:quantized}). Our SP denotes \textit{Basic search space}.}
   \label{fig:images}
\end{figure*}

In Figure~\ref{fig:images}, we provide with visual results for quantized and full precision models.

% \begin{figure}[ht]
%    \begin{center}
     
%     \end{center}
% \caption{SSIM ESPCN}
% \label{fig:espcn_grid_ssim}
% \end{figure}

% TODO
% 1. Initialize parameters and edge weights alpha_{ijb} to get architecture from our search space
% 2-3. For each step:
% 2.1. Add quantized noise for formulas (7) and (8)
% 2.2. Calculate the loss function for alphas using the alpha training sample
% 2.3. Run backpropagation to get derivatives for alphas 
% 3.1. Add quantized noise for formulas (7) and (8)
% 3.2. Update alphas 
% 3.3. Calculate the loss function for weights using the weights training sample
% 3.4. Run backpropagation to get derivatives for weights
% 3.5. Update weights
% 4. Select edges with the highest alphas
% 5. Retrain the architecture from scratch using the final training sample

%   \fbox{\rule{0pt}{0.5in} \rule{0.9\linewidth}{0pt}}

\begin{figure}[ht]
\center{\includegraphics[scale=0.33]{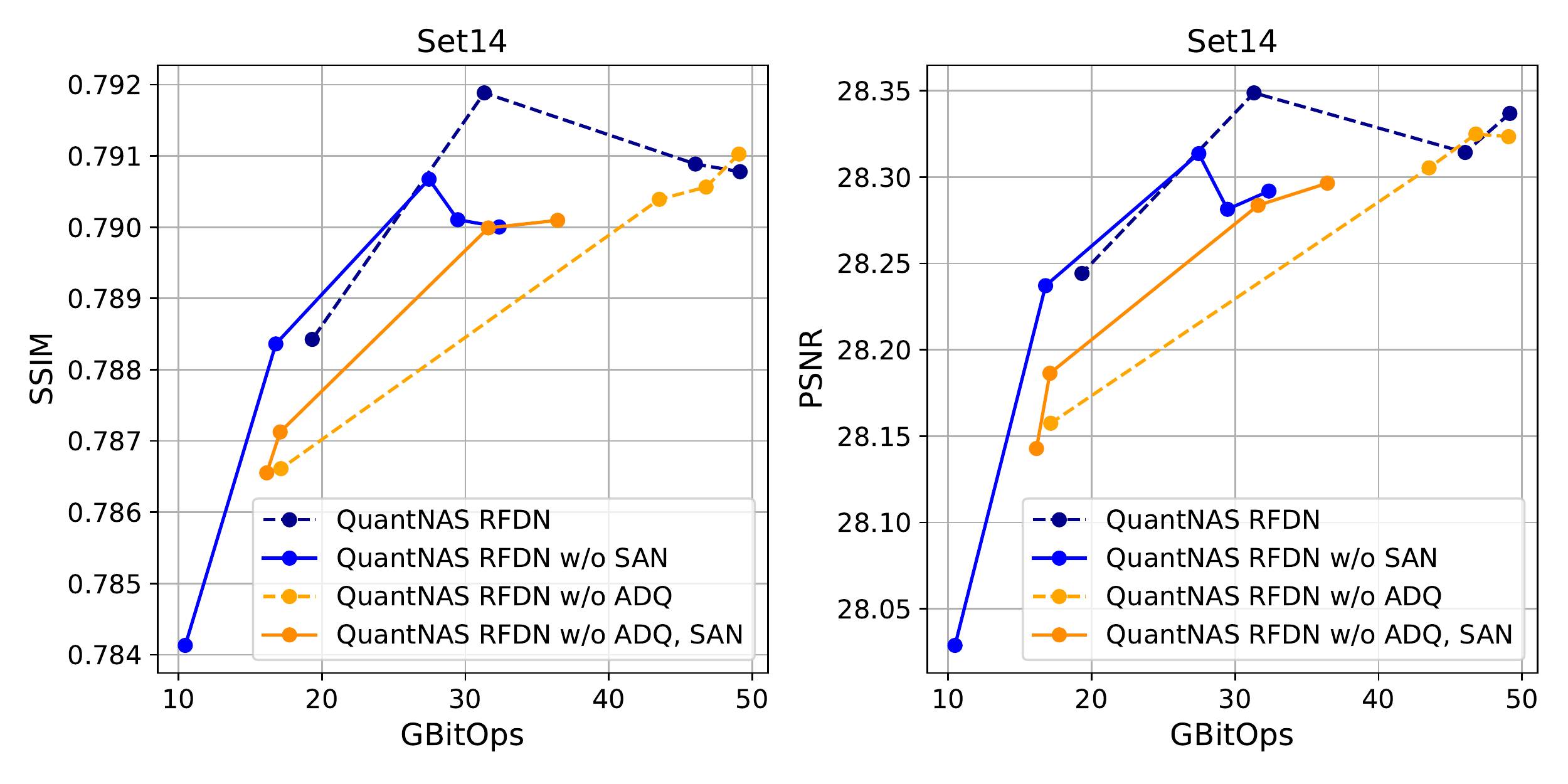}}
\caption{Comparison of results from Fig.~\ref{fig:ablation} for different metrics: SSIM and PSNR. As we can see, each metric gives a similar result.}
\label{fig:ablation_psnr}
\end{figure}

\end{document}